\documentclass[journal ]{new-aiaa}
\usepackage[utf8]{inputenc}
\usepackage{textcomp}
\usepackage{bm}
\usepackage{graphicx}
\usepackage{amsmath}
\usepackage[version=4]{mhchem}
\usepackage{siunitx}
\usepackage{longtable,tabularx}
\usepackage{lineno}
\usepackage{subcaption}
\usepackage{caption}
\usepackage{threeparttable}
\usepackage{appendix}
\usepackage{bbold}
\DeclareMathAlphabet{\mymathbb}{U}{bbold}{m}{n}
\newcommand{\eye}[1]{\mathbb{I}_{#1}}
\newcommand{\zeros}[2]{\mymathbb{0}_{#1\times#2}}
\usepackage{mathtools}
\usepackage{scalerel}

\title{High-accuracy vision-based attitude estimation system for air-bearing spacecraft simulators}

\author{Fabio Ornati, \footnote{Ph.D. Student, fabio.ornati@polimi.it.} Gianfranco Di Domenico,\footnote{Ph.D. Student, gianfranco.didomenico@polimi.it.} Paolo Panicucci,\footnote{Assistant Professor, paolo.panicucci@polimi.it.} and Francesco Topputo\footnote{Full Professor, francesco.topputo@polimi.it.}}
\affil{Department of Aerospace Science and Technology, Politecnico di Milano, Via La Masa 34, 20156, Milan, Italy}

\renewcommand{\vec}[1]{\bm{#1}}
\begin{document}

\maketitle

\begin{abstract}
Air-bearing platforms for simulating the rotational dynamics of satellites require highly precise ground truth systems. Unfortunately, commercial motion capture systems used for this scope are complex and expensive. This paper shows a novel and versatile method to compute the attitude of rotational air-bearing platforms using a monocular camera and sets of fiducial markers. The work proposes a geometry-based iterative algorithm that is significantly more accurate than other literature methods that involve the solution of the Perspective-n-Point problem. Additionally, auto-calibration procedures to perform a preliminary estimation of the system parameters are shown. The developed methodology is deployed onto a Raspberry Pi 4 micro-computer and tested with a set of LED markers. Data obtained with this setup are compared against computer simulations of the same system to understand and validate the attitude estimation performances. Simulation results show expected 1-sigma accuracies in the order of $\sim$ 12 arcsec and $\sim$ 37 arcsec for about- and cross-boresight rotations of the platform, and average latency times of 6 ms.
\end{abstract}

\section*{Nomenclature}

{\renewcommand\arraystretch{1.0}
\noindent\begin{longtable*}{@{}l @{\quad=\quad} l@{}}
\multicolumn{2}{@{}l}{\textbf{Reference frames:}}\\
$\mathcal{B}$  & body frame (3D) \\
$\mathcal{N}$  & inertial frame (3D)\\
$\mathcal{C}$  & camera frame (3D)\\
$\mathbb{C}$ & image frame (2D)\\
\multicolumn{2}{@{}l}{\textbf{Variables and indexes:}}\\
$N_m$ & number of marker points\\
$i$ & marker index\\
$N_i$ & number of images\\
$j$ & image index\\
$N_p$ & number of additional patterns\\
$k$ & additional patterns index\\
\multicolumn{2}{@{}l}{\textbf{3D geometrical quantities:}}\\
${^{\mathcal{C}}\vec{r}_{n/c}}$ & coordinates of center of rotation in camera frame \\
$^{\mathcal{B}}\vec{r}_{b/n}$ & vector from center of rotation to body frame origin expressed in body frame\\
${^{\mathcal{B}}\vec{r}_{s_k/b}}$ & coordinates of $k$-th additional pattern frame origin expressed in body frame \\
${^{\mathcal{B}}\vec{r}_{i/b}}$ & coordinates of $i$-th marker in body frame\\
${^{\mathcal{N}}\vec{r}_{i/n}}$ & coordinates of $i$-th marker in inertial frame\\
${^{\mathcal{C}}\vec{r}_{i/c}}$ & coordinates of $i$-th marker in camera frame\\
\multicolumn{2}{@{}l}{\textbf{Vectorial quantities in image frame:}}\\
$^{\scaleto{\mathbb{C}}{3.7pt}}\vec{R}^{h}_i$ &  undistorted pixel coordinates of $i$-th marker expressed with homogeneous form \\
$^{\scaleto{\mathbb{C}}{3.7pt}}\vec{R}^{u}_i$ & undistorted pixel coordinates of $i$-th marker\\
$^{\scaleto{\mathbb{C}}{3.7pt}}\vec{R}^{n,u}_i$ & undistorted normalized pixel coordinates of $i$-th marker\\
$^{\scaleto{\mathbb{C}}{3.7pt}}\vec{R}^{n}_i$ & normalized pixel coordinates of $i$-th marker\\
$^{\scaleto{\mathbb{C}}{3.7pt}}\vec{R}^{n}_i$ & pixel coordinates of $i$-th marker\\
\multicolumn{2}{@{}l}{\textbf{Rotations:}}\\
$[CN]$ & rotation matrix from inertial frame to camera frame \\
$[BS_k]$ & rotation matrix from $k$-th additional pattern frame to body frame\\
$[NB]$ & rotation matrix from body frame to inertial frame\\
$\vec{q}_{_{AB}}$ & unit rotation quaternion representing the rotation matrix $[AB]$\\
$\vec{p}_{_{AB}}$ & vector of mininal parameters representing the rotation quaternion $\vec{q}_{_{AB}}$\\
\multicolumn{2}{@{}l}{\textbf{Optical parameters:}}\\
$K$ & intrinsic camera matrix\\
$c_x$ & x coordinate of the principal point\\
$c_y$ & y coordinate of the principal point\\
$f_x$ & camera focal length for the x-axis\\
$f_y$ & camera focal length for the y-axis\\
$s$ & axial skew\\
$w_{p}$ & $p$-th radial distortion coefficient\\
\end{longtable*}

\section{Introduction}

Hardware-In-the-Loop (HIL) testing is a powerful framework to assess the functionality, reliability, and performance of space-borne systems. 
In HIL tests, the hardware components are subject to the same kind of stimuli that they would face during their on-orbit lifetime. In particular, the relevant environment, dynamics, and scenarios are recreated on-ground both physically and through computer simulations.
This allows the identification of potential issues and unforeseen interactions between the developed hardware and software before their final deployment in space. 

HIL simulations are particularly indicated for critical spacecraft subsystems such as the Attitude and Orbit Control System (AOCS). Since the beginning of the space race, AOCS hardware components such as attitude sensors, reaction wheels, thrusters, and control algorithms and policies have been tested using air bearings simulators \cite{menon2007issues,schwartz2003historical}. These kinds of hardware equipment are constituted by moving platforms sustained by a thin film of air. The flow of air acts as a cushion, preventing high-friction contacts between the platform and the sustaining surface. This allows the reproduction of the spacecraft's free-body motion in multiple Degrees of Freedom (DoF). 
In particular, spherical bearings are used to emulate the 3-DoF rotational dynamics of the spacecraft. A historical review of these kinds of test beds can be found in \citet{schwartz2003historical}. More recent examples can be found in \cite{chesi2015dynamic,schwartz2004system,kim2009automatic,prado2005three,modenini2020dynamic,wu2014low}.
Linear air bearings, instead, allow for planar movements in two directions on a flat table \cite{ReviewPlanar}. These facilities are generally used to simulate proximity operations and formation flight with multiple spacecraft \cite{zappulla2017dynamic,sabatini2012design,nieto2021concurrent}.
By combining linear and spherical air bearings it is possible to increase the number of DoFs that can be emulated by the facility \cite{chesi2015dynamic,cho20095,saulnier2014six,nakka2018six}. 

A fundamental aspect of air-bearing facilities is the external metrological system used to determine the actual state of the platform under testing. This kind of Ground Truth System (GTS) provides readings of the platform position and/or orientation, which serve as a baseline to compare the accuracy and precision of the deployed AOCS subsystem. 
Generally, it is required that the GTS shall attain a level of precision at least one order of magnitude smaller than the estimation lower bound of the system under testing.  Moreover, its readings should be acquired with a high sampling rate to characterize high-frequency dynamical motions of the platform.

Most of the GTS that have been employed are vision-based. In fact, unlike other kinds of metrological solutions such as indoor GPS (iGPS) and Inertial Measurement Units (IMUs), vision-based systems are not subject to dynamical effects and electro-magnetic inferences \cite{wang2011experimental,mautz2012indoor}. 
Some of the facilities presented in the literature use commercial motion-capture solutions, such as the Vicon®\footnote{Vicon® web-page: \href{https://www.vicon.com/} {\nolinkurl{www.vicon.com}} last accessed on Sept. 2023.} system, as GTS \cite{zappulla2017dynamic}. These kinds of solutions use a large number of synchronized cameras to detect and compute the 3-dimensional position of a set of markers applied to the platform. These systems are particularly expensive, with total costs for a complete setup, including hardware, installation, and commissioning, exceeding tens of k€. Additionally, they do not allow full control of the estimation procedure as it is carried out by proprietary software.

Cheaper alternative solutions have been discussed by several authors \cite{wu2014low,modenini2020dynamic,hua2015new,bing2016spacecraft}. All these systems are based on cameras that detect fiducial markers installed onto the simulator platform. The fiducial markers consist of either checkerboard patterns or led arrays. 
Checkerboard patterns, used in the setup of \cite{modenini2020dynamic}, can be easily printed with the required geometrical precision and can be detected by well-known corner detection techniques \cite{geiger2012automatic}. However, since they are not actively illuminated, they require relatively long exposure times which can cause motion-blurring effects when the platform is in motion.
LED marker arrays, such as those used in the work of \cite{wu2014low}, are more difficult to manufacture since they require precise placement of the electronic components. However, since their brightness is much higher than that of the surrounding ambient, images with very brief exposure times can be acquired. In those images, only the LEDs spots are visible against a dark background. This allows the use of centroiding algorithms for detecting the LED positions, which can be both simpler and more accurate than the techniques used for checkerboard patterns \cite{vyas2009performance}.

All the cited works (\cite{wu2014low,modenini2020dynamic,hua2015new,bing2016spacecraft}) use the marker points to compute the attitude and position of the air-bearing platform by solving the Perspective-$n$-Point (P$n$P) problem. The P$n$P is the problem of determining the pose of a camera given the corresponding position of a known 3-D set of points in an image \cite{lepetit2009ep}. The solution requires the knowledge of the camera's intrinsic parameters, which can be estimated through widely-used calibration procedures \cite{zhang2000flexible}.
It is important to note that solving the P$n$P yields both a rotation matrix and a translation vector between the camera and the platform frames. If the platform has one or more translational DoFs, both quantities contribute to determining the current state of the platform. However, in the case of 3-DoF rotational simulators, the rotation and the translation of the pattern frames are not independent, since all the fiducial marker points are constrained to move around a fixed center of rotation.  

Previous works which involved rotational platforms do not take advantage of this constraint. In fact, in the algorithms that they propose, the translation vector retrieved through the solution of the P$n$P is not used and it is simply discarded. In those cases, six independent parameters are estimated by the P$n$P despite just three DoFs being available. 
This leads to sub-optimal attitude solutions, which are less accurate than those that would be obtained if only rotations were estimated.  

This paper tackles the problem of determining the attitude of 3-DoF rotational air-bearing platforms by proposing a novel methodology that exploits the fixed geometric architecture of the system. In the developed method, the geometrical constraints are naturally embedded, leading to the estimation of a minimal set of three independent parameters that describe the rotation of the platform.

The following of the paper is described as follows: 
Section \ref{sec:2} discusses the method, focusing on the theoretical model and the algorithms for performing the estimation of the attitude and the geometrical characteristics of the system. Section \ref{sec:3} shows a prototypal hardware implementation of the method that uses a set of fiducial LED markers and an industrial computer vision camera. Obtained results are compared with data gathered through Monte Carlo simulations to determine the expected system accuracy and the improvement in performance with respect to the current state-of-the-art P$n$P-based methods.
Finally, conclusions and possible future work are presented in Section \ref{sec:4}.

\section{Theoretical framework}
\label{sec:2}
\subsection{Notation}
The following notation is used in this document:
\begin{itemize}
\item Two- and three-dimensional vectors are denoted, respectively, with upper- and lowercase bold text, such as $\vec{R}$ and $\vec{r}$.
\item Vector initialization is performed with parentheses, such as  $\vec{a} = \left(b,\; c,\;d\right)^\top$.
\item Matrix initialization is performed with square brackets, such as  $A = \left[\vec{a},\; \vec{b},\;\vec{c}\right]$.
\item $\mathcal{A}= \{a,\, \vec{a}_1,\, \vec{a}_2,\, \vec{a}_3\}$ is the 3D reference frame centered in the 3D point $a$ with unit axes $\vec{a}_1$,  $\vec{a}_2$, and  $\vec{a}_3$. All the reference frames are right-handed and orthonormal.
\item Rotation matrices are in plain text in brackets. The  rotation matrix from $\mathcal{A}$ to  $\mathcal{B}$ is $[BA]$. All rotations have a passive function. 
\item The unit quaternion for the rotation $[AB]$ is $\vec{q}_{_{AB}}$. Quaternions follow the scalar-first convention, so that $\vec{q} = (q_w,\,\vec{q}_v)^\top$.
\item The 3D vector from point $p$ to point $q$, expressed in frame $\mathcal{A}$ is denoted $^\mathcal{A}\vec{r}_{q/p}$. 
\item $\mathbb{A}= \{A,\, \vec{A}_1,\, \vec{A}_2\} $ is the 2D reference frame centered in $A$ with orthogonal unit axes $\vec{A}_1$ and $\vec{A}_2$.
\item The 2D vector from the origin of frame $\mathbb{A}$ to point $p$, expressed in frame $\mathbb{A}$ is denoted $^{\mathbb{A}}\vec{R}_{_{p}}$
\item The 3D vector $\vec r$ in homogeneous form is labeled $\vec{r}^{h}$ and the 2D vector $\vec{R}$ in homogeneous
form is labeled $\vec{R}^{h}$.
\item The identity matrix of dimension $n$ is labeled $\eye{n}$.
\item The $n$-by-$m$ matrix of zeros is labeled $\zeros{n}{m}$.
\end{itemize}

\subsection{Geometry and frames}
\label{ssec:geometry}
The proposed system architecture comprises three main elements: a spherical air-bearing platform, a set of fiducial markers, and an observing camera.
The air bearing allows quasi-frictionless three-axial rotations around a fixed Center of Rotation (CoR). A set of fiducial markers are fixed at the top of the platform and pivot around the CoR during movements. Finally, a monocular camera is placed above and is able to detect the markers. 

\begin{figure}[b!]
\centering
\includegraphics[width=.65\textwidth]{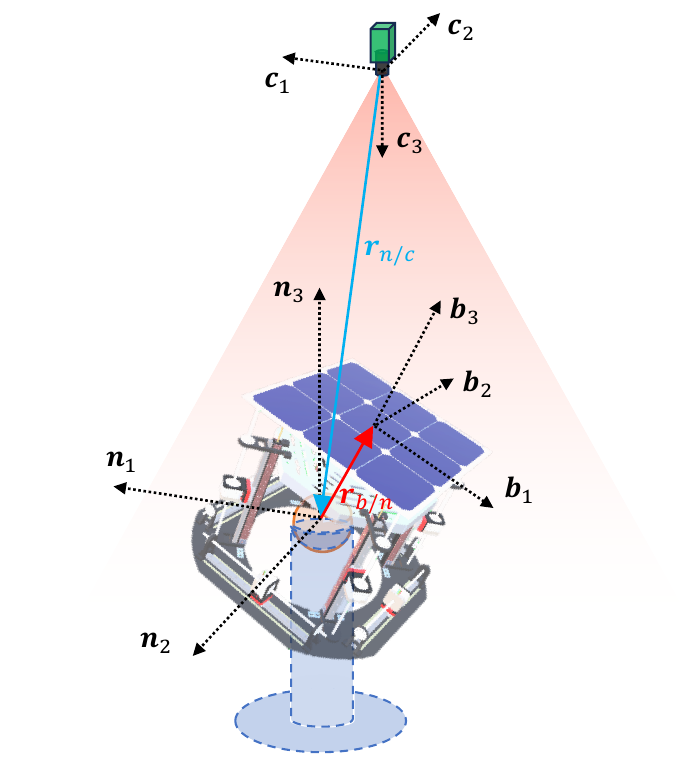}
\caption{Scheme of ground truth system geometry and frames.}
\label{fig:geometry_scheme}
\end{figure}

Each of these elements is associated with a 3-D coordinate frame, as shown in Fig. \ref{fig:geometry_scheme}. 
Let $\mathcal{N}$ and $\mathcal{C}$ be the inertial and camera reference frame, respectively. The former has its origin in the CoR and the latter at the principal point of the camera objective.
The vector $^{\mathcal{C}}\vec{r}_{n/c}$ relates the positions of the origins of the two frames. Note that the frames are defined such that the rotation matrix $[CN]$ from $\mathcal{N}$ to $\mathcal{C}$ is fixed as:

\begin{equation}
    [CN] = \begin{bmatrix} 
    1 & 0 & 0 \\ 0 & -1 & 0 \\ 0 & 0 & -1
    \end{bmatrix}
\end{equation}

Let $\mathcal{B}$ be the body-fixed reference frame that rotates coherently with the platform. This is the frame in which the coordinates of the fiducial marker points are expressed. The origin of $\mathcal{B}$ is related to the origin of $\mathcal{N}$ by the vector $^{\mathcal{B}}\vec{r}_{b/n}$. 
Note that the complete set of fiducial markers may be composed of more than one individual pattern. 
If this is the case, then the first pattern defines the $\mathcal{B}$ frame, and each $k$-th additional pattern is associated with its own reference frame $\mathcal{S}_k$. The $k$-th additional frame $\mathcal{S}_k$ is related to $\mathcal{B}$ by the vector ${^{\mathcal{B}}\vec{r}_{s_k/b}}$ and the rotation matrix $[BS_k]$.

The goal of the GTS is to compute the rotation quaternion $\vec{q}_{_{NB}}$, which determines the rotation from $\mathcal{B}$ to $\mathcal{N}$ and thus the relative attitude of the air bearing platform.

\subsection{Camera projection}
\label{Camera projection}
When the camera acquires images of the fiducial pattern, each marker point is projected onto the image frame $\mathbb{C}$. The pixel coordinates of the $i$-th marker in the image frame $^{\scaleto{\mathbb{C}}{3.7pt}}\vec{R}_i$ are uniquely determined by three factors: the geometric configuration of the system, the current attitude of the platform, and the camera optical intrinsic parameters.

Let ${^{\mathcal{B}}\vec{r}_{i/b}}$ be the coordinates of the $i$-th marker in the body frame. If the marker belongs to the $k$-th additional pattern, then its position in the ${\mathcal{B}}$ frame can be computed as: 

\begin{equation}
    {^{\mathcal{B}}\vec{r}_{i/b}} = {^{\mathcal{B}}\vec{r}_{s_k/b}} + [BS_k]\; {^{{\mathcal{S}^k}}\vec{r}_{i/s_k}}
    \label{eq:pr_1}
\end{equation}

Each vector ${^{\mathcal{B}}\vec{r}_{i/b}}$ can be expressed in the inertial frame knowing the rotation attitude of the platform $[NB]$:

\begin{equation}
\label{eq:r_nin}
    {^{\mathcal{N}}\vec{r}_{i/n}}= [NB]\left({^{\mathcal{B}}\vec{r}_{i/b}}+{^{\mathcal{B}}\vec{r}_{b/n}}\right)
\end{equation}

For points in the inertial frame, the following roto-translation applies:

\begin{equation}
    ^{\mathcal{C}}\vec{r}_{i/c}= {^{\mathcal{C}}\vec{r}_{n/c}} + [CN]\; {^{\mathcal{N}}\vec{r}_{i/n}}
\end{equation} 

The projection of the marker points onto the image plane is modeled using the pinhole camera model \cite{Hartley_Zisserman_CV}. In particular, $^{\scaleto{\mathbb{C}}{3.7pt}}\vec{R}^{h}_i$, which is the homogeneous projection of $^{\mathcal{C}}\vec{r}_{i/c}$ in the image frame is described by: 

\begin{equation}
    ^{\scaleto{\mathbb{C}}{3.7pt}}\vec{R}^{h}_i = K \; {^{\mathcal{C}}\vec{r}}_{i/c}
\end{equation}

Where $K$ is the 3-by-3 intrinsic camera matrix. In its most general formulation, $K$ has 5 parameters \cite{Hartley_Zisserman_CV}: 

\begin{equation}
    K= \begin{bmatrix}
        f_x& s & c_x \\ 0 & f_y & c_y \\ 0 & 0 &1
    \end{bmatrix}
\end{equation}

where $f_x$ and $f_y$ are the focal lengths for the two axes, and $(c_x,\, c_y)$ is the location of the principal point in the image frame. The axial skew $s$ has been assumed to be always equal to zero in this work \cite{Hartley_Zisserman_CV}.

The non-homogeneous vector $^{\scaleto{\mathbb{C}}{3.7pt}}\vec{R}^{u}_i$, labeling the undistorted pixel coordinates in the image plane is obtained as:

\begin{equation}
^{\scaleto{\mathbb{C}}{3.7pt}}\vec{R}^{u}_i =  \frac{1}{ {^{\scaleto{\mathbb{C}}{3.7pt}}\vec{R}}^{h}_i \cdot \vec{C}_3}
\begin{pmatrix}
    ^{\scaleto{\mathbb{C}}{3.7pt}}\vec{R}^{h}_i \cdot \vec{C}_1  \\
    ^{\scaleto{\mathbb{C}}{3.7pt}}\vec{R}^{h}_i \cdot \vec{C}_2 
\end{pmatrix}
\end{equation}

Note that this is not the actual location of the points in the image plane, since optical distortions are present. This paper assumes radial distortions only modeled through a third-degree polynomial of the form \cite{Tang_distortionModels}:

\begin{equation}
    ^{\scaleto{\mathbb{C}}{3.7pt}}\vec{R}^{n}_i = {^{\scaleto{\mathbb{C}}{3.7pt}}\vec{R}^{n,u}_i} \left(1 + w_1 \rho_i^2 + w_2 \rho_i^4 + w_3 \rho_i^6 \right)
    \end{equation}

where:
    
    \begin{equation}
         \quad \rho_i^2 = \left({^{\scaleto{\mathbb{C}}{3.7pt}}\vec{R}^{n,u}_i}\right)^\top \left({^{\scaleto{\mathbb{C}}{3.7pt}}\vec{R}^{n,u}_i}\right)
    \end{equation}

$^{\scaleto{\mathbb{C}}{3.7pt}}\vec{R}^{n,u}_i$ is the undistorted pixel position expressed in normalized image coordinates, obtained as:

\begin{equation}
    {^{\scaleto{\mathbb{C}}{3.7pt}}\vec{R}^{n,u}_i} = \begin{pmatrix}
        \dfrac{{^{\scaleto{\mathbb{C}}{3.7pt}}\vec{R}^{u}_i}\cdot{\vec{C}_1} -c_x}{f_x}
        \\
        \dfrac{{^{\scaleto{\mathbb{C}}{3.7pt}}\vec{R}^{u}_i}\cdot{\vec{C}_2} -c_y}{f_y}
    \end{pmatrix} 
\end{equation}

Finally, the distorted pixel coordinates are computed as:

\begin{equation}
     ^{\scaleto{\mathbb{C}}{3.7pt}}\vec{R}_i = \begin{pmatrix}
        f_x\; {^{\scaleto{\mathbb{C}}{3.7pt}}\vec{R}^{n}_i}\cdot{\vec{C}_1}  
        \\
        f_y\; {^{\scaleto{\mathbb{C}}{3.7pt}}\vec{R}^{n}_i}\cdot{\vec{C}_2}
    \end{pmatrix} +\begin{pmatrix}
        c_x \\ c_y
    \end{pmatrix}
    \label{eq:pr_end}
\end{equation}

The previous equations allow to predict the coordinates of each marker in the images for a given system geometry, platform attitude, and set of camera parameters. In real applications, however, the true marker coordinates detected by the camera ${^{\scaleto{\mathbb{C}}{3.7pt}}\hat{\vec{R}}_i}$ differ from the predicted ones ${^{\scaleto{\mathbb{C}}{3.7pt}}\vec{R}_i}$. Image noise is among the main causes of this. Noise affects directly the images, degrading the precision with which the pixel coordinates can be determined. In addition, errors in the knowledge of the exact marker position may lead to imprecise and biased predictions. These errors might be caused by inaccuracies in the manufacturing or installation of the physical markers and by other mechanical effects, including deformations of the platform due to loads and thermal stresses.

\subsection{Attitude estimation}
\label{Attitude estimation}

Consider the system described in Section \ref{ssec:geometry}. Having the knowledge of all the geometrical quantities and the camera's intrinsic parameters, it is possible to compute the projection of each marker point for a given platform attitude $\vec{q}_{_{NB}}$ through equations \eqref{eq:pr_1} to \eqref{eq:pr_end}.
By subtracting the projected coordinates ${^{\scaleto{\mathbb{C}}{4pt}}{\vec{R}}_i}$ and the real ones ${^{\scaleto{\mathbb{C}}{3.7pt}}\hat{\vec{R}}_i}$ retrieved by the camera, a vector of reprojection errors $\boldsymbol{\varepsilon}$ can be identified. This vector has the size of $2N_m$-by-1, where $N_m$ is the number of markers:

\begin{equation}
    \boldsymbol{\varepsilon} = \begin{pmatrix}
        {^{\scaleto{\mathbb{C}}{3.7pt}}\vec{R}_1} - {^{\scaleto{\mathbb{C}}{3.7pt}}\hat{\vec{R}}_1} \\
        {^{\scaleto{\mathbb{C}}{3.7pt}}\vec{R}_2} - {^{\scaleto{\mathbb{C}}{3.7pt}}\hat{\vec{R}}_2} \\
        \vdots \\
        {^{\scaleto{\mathbb{C}}{3.7pt}}\vec{R}_{_{N_m}}} - {^{\scaleto{\mathbb{C}}{3.7pt}}\hat{\vec{R}}_{_{N_m}}} 
    \end{pmatrix}
\end{equation}

Supposing that all the geometrical quantities and the intrinsic parameters are precisely known, it is possible to estimate the platform attitude by finding the optimal quaternion solution $\vec{q}_{_{NB}}$ that minimizes the total quadratic reprojection error off all the marker points. The problem can be rigorously stated as:

\begin{equation}
    \min_{\vec{q}_{_{NB}}} \; \boldsymbol{\varepsilon}^\top \boldsymbol{\varepsilon} \quad \textrm{such that:} \quad \|\vec{q}_{_{NB}}\| = 1
\end{equation}

Using the quaternion components as optimization variables improves the numerical stability and avoids the evaluation of trigonometric functions during the computations. Moreover, quaternions are not subject to the numerical gimbal locking effects typical of Euler angles \cite{schmidt2001using,terzakis2014quaternion}. 
The main disadvantage of quaternions is the additional nonlinear equality constraint needed to enforce the unitary norm. This is avoided by using an alternative minimal parametrization in which the constraint is naturally satisfied. The one presented in \cite{terzakis2014quaternion} is employed for this scope. In particular, a 4-dimensional unitary quaternion $\vec{q}$ can be parametrized with a 3-dimensional vector of parameters $\vec{p} = (p_1,\, p_2, \, p_3)^\top$ as:

\begin{equation}
    \vec{q} = \frac{2}{\alpha^2+1}\left(p_1,\,p_2,\,p_3,\,\frac{1-\alpha^2}{2}  \right)^\top 
\end{equation}

where:

\begin{equation}
    \alpha^2=p_1^2+p_2^2+p_3^2
\end{equation}

Using this parametrization, the problem is simplified as: 

\begin{equation}
    \min_{\vec{p}_{_{NB}}} \; \boldsymbol{\varepsilon}^\top \boldsymbol{\varepsilon}
\end{equation}

where $\vec{p}_{_{NB}}$ is the vector of parameters that describe the rotation $[NB]$.

The solution to the problem can be obtained through an iterative least-squares method. At each iteration step $k$ the solution is updated as:

\begin{equation}
 \vec{p}_{_{NB}}^{(k+1)} = \vec{p}_{_{NB}}^{(k)} - \left({J_{p}^{(k)}}^\top J_{p}^{(k)} \right)^{-1} {J_{p}^{(k)}}^\top \boldsymbol{\varepsilon}^{(k)} 
\end{equation}

where $J_{p}^{(k)} = \dfrac{ \partial \boldsymbol{\varepsilon}^{(k)} }{ \partial \vec{p}_{_{NB}}^{(k)}}$ is the $2N_m$-by-3 Jacobian matrix of $\boldsymbol{\varepsilon}$ at the step $k$. The detailed formulation of $J_{p}$ is reported in \hyperref[sec:appA]{Appendix A}.

\subsection{System calibration}
\label{System calibration}
The method proposed in Section \ref{Attitude estimation} requires accurate knowledge of the camera characteristics and the geometrical parameters of the system. Even small errors in those could lead to biased and inexact attitude solutions. 

Generally, the camera intrinsic parameters can be estimated through camera calibration procedures, which rely on the acquisition of several images of calibration patterns \cite{zhang2000flexible}. The geometric properties, instead, can only be determined through manual measurements or with accurate CAD models of the entire facility. These measurements can be especially troublesome to acquire with the necessary precision since they involve three-dimensional vectorial quantities. The methodology proposed in this paper exploits the rotating platform itself as a calibration device to perform a batch estimation of all the geometric and optical parameters. 

In particular, the optical parameters are seven: the four projection matrix parameters ($f_x$, $f_y$, $c_x$, and $c_y$), and the three polynomial distortion coefficients $w_{1,2,3}$. These are arranged in two vectors: $\vec{k} = (f_x,\;f_y,\;c_x,\;c_y)^\top$ and $\vec{w} = (w_1,\;w_2,\;w_3)^\top$.
The geometrical parameters considered are the two 3-D vectors $^{\mathcal{B}}\vec{r}_{b/n}$ and $^{\mathcal{C}}\vec{r}_{n/c}$, and the translation ${^{\mathcal{B}}\vec{r}_{s_k/b}}$ and rotation $\vec{q}_{_{BS_k}}$ of each additional pattern with respect to the body frame. For the same reasons discussed in Section \ref{Attitude estimation}, each rotation quaternion $\vec{q}_{_{BS_k}}$ has been parametrized with an equivalent minimal vector of parameters $\vec{p}_{_{BS_k}}$.

The batch estimation of this large set of parameters requires a vast amount of data, which can be retrieved by acquiring several sets of images of the platform whilst in different attitude positions. In particular, if $N_i$ images of $N_m$ markers are acquired, the total number of available scalar measurements, which are the coordinates of the markers in the images, is $m = 2 N_i N_m$. Instead, if $N_p$ is the number of additional patterns, there is a total of $ p = 13 + 6N_p + 3 N_i$ scalar parameters to be estimated. Seven of these are optical parameters, $6 + 6N_p$ are those related to the fixed geometry, and $3 N_i$ determine the attitude of the platform $[NB]_j$ in the $j$-th image.
A least-squares estimation of the parameters can be performed as long as $m \geq p$. 
Let $\vec{v}$ be the $p$-by-1 vector containing all the parameters and $\boldsymbol{\eta}$ the $m$-by-1 vector of reprojection errors for all the images: 

\begin{equation}
    \vec{v} = \left(\vec{k},\; \vec{w},\; {^{\mathcal{B}}\vec{r}_{b/n},\; {^{\mathcal{C}}\vec{r}_{n/c}},\; {^{\mathcal{B}}\vec{r}_{s_1/b}},\; \vec{p}_{_{BS_1}},\; \dots,\; {^{\mathcal{B}}\vec{r}_{s_{N_p}/b}},\; \vec{p}_{_{BS_{N_p}}},\; \vec{p}_{_{NB_{1}}},\; \dots,\; \vec{p}_{_{NB_{N_i}}}  }  \right)^\top
\end{equation}

\begin{equation}
\boldsymbol{\eta} = \left({\boldsymbol{\varepsilon}_1},\;{\boldsymbol{\varepsilon}_2}, \;\dots,\; {\boldsymbol{\varepsilon}_{N_i}}  \right)^\top
\end{equation}

Then, by labelling as $J^{(k)} = \dfrac{ \partial \boldsymbol{\eta}^{(k)} }{ \partial \vec{v}^{(k)}}$ the $m$-by-$p$ Jacobian matrix of $\boldsymbol{\varepsilon}$ at the iteration step $k$, the solution at the step $k+1$ can be computed as: 

\begin{equation}
 \vec{v}^{(k+1)} = \vec{v}^{(k)} - \left({J^{(k)}}^\top J^{(k)} \right)^{-1} {J^{(k)}}^\top \boldsymbol{\eta}^{(k)} 
\end{equation}

All the partial derivatives that constitute the Jacobian $J^{(k)}$ are explicitly reported in \hyperref[sec:appB]{Appendix B}.

\subsection{Methodology workflow}
\label{Methodology workflow}
The proposed methodology is divided into two phases: offline calibration and online attitude estimation. The complete workflow is shown in Fig. \ref{fig:workflow_scheme}. During the calibration phase, the system parameters are estimated using the algorithm of Section \ref{System calibration}. The estimation algorithm requires an initial guess $\vec{v}^{(0)}$. The geometrical parameters of $\vec{v}^{(0)}$ are initialized through coarse manual measurements of the system. The optical properties are set to the theoretical values stated by the camera and objective manufacturers.

\begin{figure}
\centering
\includegraphics[width=0.8\textwidth]{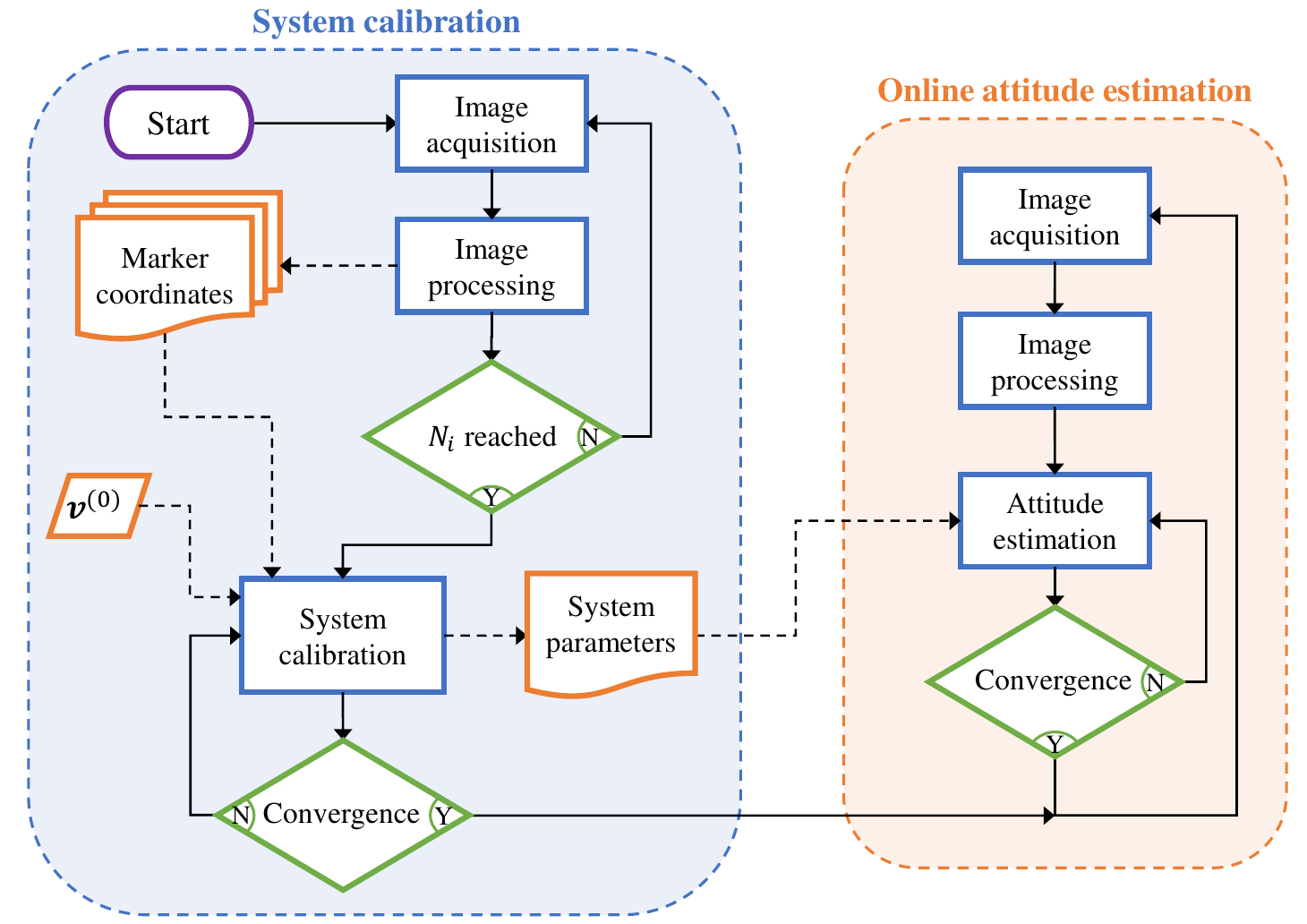}
\caption{Ground truth operational workflow. Dashed lines indicate the data workflow.}
\label{fig:workflow_scheme}
\end{figure}

While the calibration images are being acquired, the platform is rotated by an operator or by its internal attitude control system. The initial guess for the attitude of the platform in each calibration image $\vec{p}_{_{NB_{j}}}$ is preliminarily determined using the P3P problem solver \cite{gao2003complete} implemented in MATLAB with the initial camera parameters. 
Note that, if $N_i$ is large, the calibration process can become computationally burdensome. In particular, each iteration step requires the computation of the $m$-by-$p$ Jacobian and the inversion of a $p$-by-$p$ matrix.

Once the calibration phase reaches convergence, it is possible to use the newly estimated parameters to perform the online attitude estimation in continuous mode. At each cycle, a new image is acquired and directly processed to determine the coordinates of the marker points. Then, using the iterative algorithm discussed in Section \ref{Attitude estimation}, an attitude solution is found. Since only three scalar variables are estimated, the process can be run quickly to provide a low-latency attitude solution. Moreover, if the overall refresh rate is sufficiently high or the platform rotates slowly, it is possible to use the solution of the previous cycle as initial guess, considerably reducing the the processing time and the number of iterations needed to achieve convergence. 

\section{Hardware implementation}
\label{sec:3}
\subsection{Hardware description}
STASIS (SpacecrafT Attitude SImulation
System) is a 3-DoF air-bearing simulation platform currently under development at the DART (Deep-Space Astrodynamics Research and Technology) group of Politecnico di Milano \cite{di2022stasis}. STASIS is one of the key elements of the EXTREMA Simulation Hub (ESH) \cite{orbitalSimulationHub}, the integrated HIL simulator facility of the European Research Council -funded EXTREMA\footnote{EXTREMA project web-page: \href{https://dart.polimi.it/extrema-erc/} {\nolinkurl{https://dart.polimi.it/extrema-erc/}}, last accessed on Sept. 2023.} (Engineering eXTremely Rare Events in astrodynamics for deep-space Missions in Autonomy) project \cite{di2022erc_EXTREMAchapter,Extrema_IAC2021}. 

The ESH is used to perform dynamical HIL simulations of the Guidance, Navigation, and Control sub-systems of an autonomous deep-space CubeSat during its interplanetary transfer leg. Within this framework, STASIS reproduces the rotational dynamics of the spacecraft and the control applied by the onboard actuators.
The GTS developed for STASIS provides instantaneous readings of the platform's true attitude. These measurements have multiple and important uses during the simulations. In particular, in the initial setup phase, they are used by the balancing algorithm to correct the position of movable masses onboard. Similarly, the GTS is used to estimate the inertial moment characteristics of the platform. 
Then, the attitude readings are used to propagate the true spacecraft state in accordance with the actual thrusting direction. This task is performed by the SPESI simulator \cite{giordano2023spesi}, which uses the real values of thrust measured from the ETHILE facility \cite{morselli2022ethile}.
Additionally, the GTS readings are used to generate realistic night-sky scenes of the celestial sphere as they would be seen by the spacecraft. The scenes are corrected for misalignment and distortions  \cite{pugliatti2022tinyv3rse,panicucci2022tinyv3rse} and are displayed in the RETINA optical facility \cite{andreis2023towards}. The images acquired by the camera are then used to perform the internal attitude determination and to test celestial navigation techniques.
Thanks to the precise reference provided by the GTS, it is possible to validate the performance and the accuracy of both the attitude determination and attitude control processes.

STASIS, which can be seen in Fig. \ref{fig:stasis_image}, has bounding-box dimensions of 55 by 55 by 35 cm. Yaw rotations (around the $\vec{b}_3$ axis) are unconstrained, while pitch and roll movements ($\vec{b}_2$ and $\vec{b}_1$ axes, respectively) are limited to	$\pm$ 22 degrees due to mechanical constraints. 

\begin{figure}[t]
\centering
\includegraphics[width=0.55\textwidth]{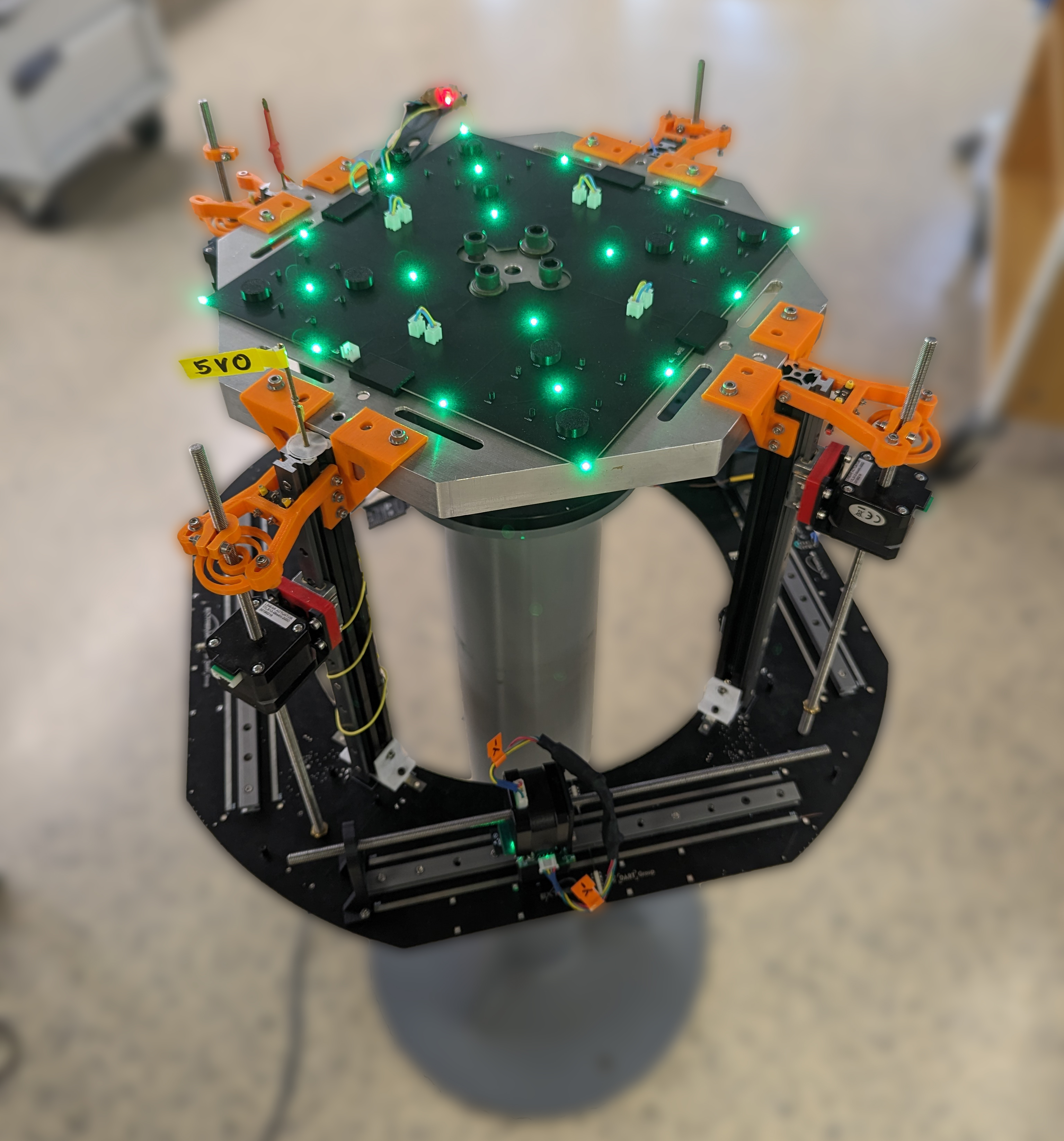}
\caption{STASIS platform at its current development state.}
\label{fig:stasis_image}
\end{figure}
The camera selected for the GTS, a FLIR BFS-U3-31S4M-C\footnote{FLIR Blackfly-s-usb3 web-page: \href{https://www.flir.com/products/blackfly-s-usb3/} {\nolinkurl{https://www.flir.com/products/blackfly-s-usb3/}}, last accessed on Sept. 2023.}, is an industrial-grade camera for computer-vision applications. The imaging sensor has a resolution of 1536 x 2048 pixels in a 1/1.8" format. The focal length of the objective is 12 mm, yielding a Field-of-View (FoV) of 32.8 by 24.9 degrees. Since the camera is placed 1.1 m above the patterns, the FoV guarantees the observation of the entire platform in any possible attitude position.

The fiducial pattern is located on top of the platform and uses an array of active LEDs as markers. A total of 20 green LEDs are assembled onto 4 individual Printed Circuit Boards (PCBs). The architecture of the patterns and their dimensions are shown in Fig \ref{fig:pattern_scheme}. A specific LED model\footnote{LED datasheet: \href{https://www.kingbrightusa.com/images/catalog/SPEC/APG0603ZGC-5MAV.pdf} {\nolinkurl{https://www.kingbrightusa.com/images/catalog/SPEC/APG0603ZGC-5MAV.pdf}}, last accessed on Sept. 2023.} is selected for its high-luminous intensity of 280 mcd and its small dimensions of 0.65 x 0.35 mm (with a "0201" standard footprint).
The small size has two main advantages. First, it requires the assembler of the PCB to meet tighter tolerances when placing the components to guarantee their functionality. Second, it makes the LED appear more similar to a point-wise object despite its parallelepipedal shape.
The high luminance, instead, enables the acquisition of images with very low exposure timings, reducing the motion-blurring effects due to the platform motion. In particular, all the images are acquired with an exposure time of just 11 \textmu s. With this exposure time, all the pixels in the images that are not illuminated by the LEDs are extremely dark, having 8-bit digital count levels below 4. 

\begin{figure}[t]
\centering
\includegraphics[width=0.7\textwidth]{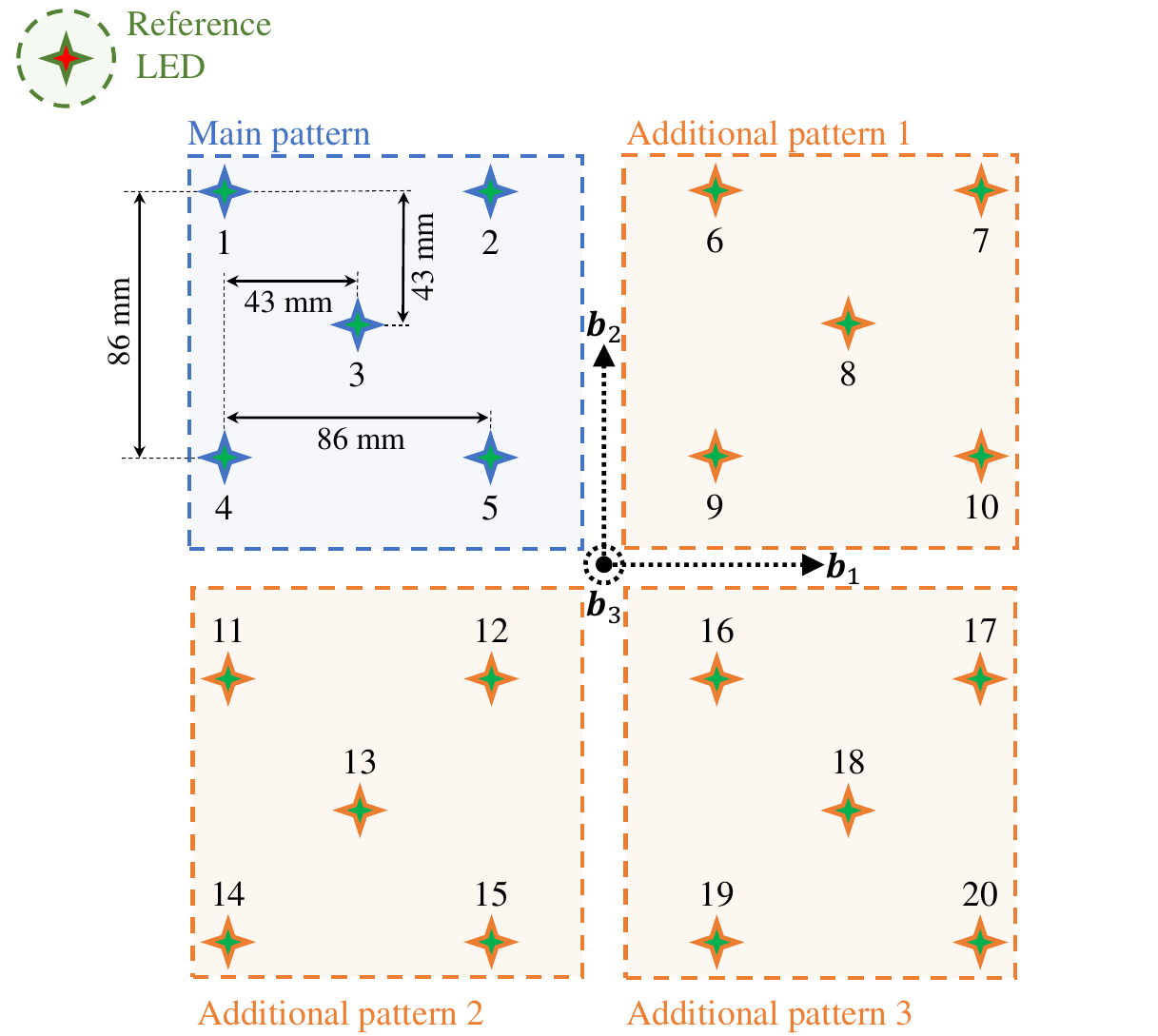}
\caption{LED patterns and geometrical dimensions.}
\label{fig:pattern_scheme}
\end{figure}

\subsection{Image processing and marker identification}
A Weighted Center of Mass (WCoM) centroiding technique is used to find the position of each LED in the images ${^{\scaleto{\mathbb{C}}{3.7pt}}\hat{\vec{R}}_i}$ \cite{vyas2009performance}. In particular, the position of the $i$-th centroid is computed as: 

\begin{equation}
    {^{\scaleto{\mathbb{C}}{3.7pt}}\hat{\vec{R}}_i} = \cfrac{1}{ \sum\limits_j^{n_i} I_{j} W_j} \sum\limits_{j=1}^{n_i} \left( I_{j} W_j {^{\scaleto{\mathbb{C}}{3.7pt}}\vec{R}_j} \right)
\end{equation}

where $n_i$ is the number of adjacent pixels of the centroid $i$ with digital count values $I_{j}$ above the threshold $I_{\rm{min}} = 5$, ${^{\scaleto{\mathbb{C}}{3.7pt}}\vec{R}_j}$ expresses the coordinates of the center of each pixel $j$, and $W_j$ is an intensity-dependent weight function. The value of $W_j$ has been set equal to $I_{j}$ as proposed by \cite{vyas2009performance}. 

Once all the centroids are computed, each one has to be assigned to the corresponding LED. The marker recognition is performed thanks to an additional LED positioned outside the main pattern at the top-left corner. As shown in Fig. \ref{fig:pattern_scheme}, since the reference LED is the farthest away from all the others, it can be easily identified by computing the distance from the mean position of all the centroids. Then, the LEDs at the minimum and maximum distance from it are identified as LED 1 and LED 20, respectively. From there, all the LEDs are first sorted depending on their distance from LED 1 in the direction of LED 20, then they are sorted again depending on their bearing angle with respect to LED 1. 
This procedure is fast as it requires very few computations and can easily be adapted for similar patterns with a higher number of LEDs.

\subsection{Calibration results}
The calibration of the system is carried out using a total of 350 images of the platform while being rotated by an operator. The system's parameters are estimated using the procedure shown in Section \ref{System calibration}.
Note that, since the 4 patterns are all co-planar, it is possible to avoid the estimation of some parameters. 
In particular, by imposing this constraint, the rotation matrices $[BS_k]$ of the three additional patterns with respect to the body frame have only one free parameter. This is because the only possible rotation is the one around the $\vec{b}_3$ axis. When third-axis rotations are expressed using the parametrization, the only parameter that changes is $p_1$, therefore $p_2$ and $p_3$ are constant and their derivatives are zero.
Similarly, since all the translation vectors ${^{\mathcal{B}}\vec{r}_{s_k/b}}$ are coplanar, the third component of each vector is constant and has null derivatives.
With these constraints in place, the number of unknown parameters $p$ is reduced from $13 + 6N_p + 3 N_i$ to $13 + 3N_p + 3 N_i$. For the considered case, since $N_p = 3$ and $N_i = 350$, a total of $p = 1072$ variables are estimated during the calibration procedure. 

The calibration algorithm usually converges in 4-6 iterations starting from the initial guess generated through manual measurements.
Figure \ref{fig:2DPDF_reprErr} shows the 2-D distribution of the components of the reprojection errors after a typical calibration run. The value of the standard deviation of the errors is $\sigma_{\mathrm{err}} = 0.115$ pixel. The mean absolute error, instead, is 0.138 pixel, while the total residual of the calibration is $r^2 = \boldsymbol{\eta}^\top \boldsymbol{\eta}= 173.1$ pixel$^2$.
\begin{figure}[t]
\centering
\includegraphics[width=0.6\textwidth]{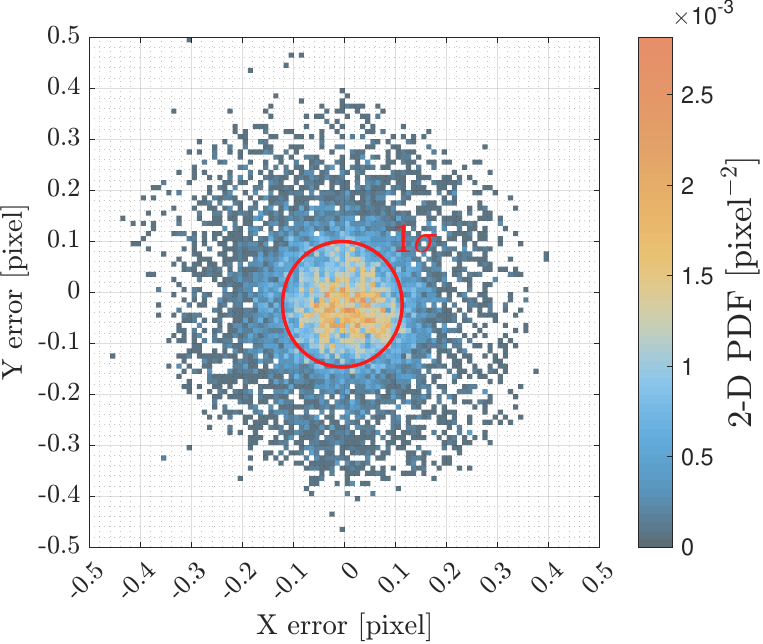}
\caption{2-D PDF of reprojection errors after calibration.}
\label{fig:2DPDF_reprErr}
\end{figure}

\begin{table}
\centering
\caption{\label{tab:estimated_uncertainty}  Value of estimated parameters and related uncertainty.}
\begin{tabular}{lccc}
\hline
Parameter & Unit & Estimated value & 1-$\sigma$ uncertainty \\ \hline
$(f_x,\;f_y)$ & pixel & $(3481.8,\;3479.6)^\top$ & $(0.957,\;0.960)^\top$ \\
$(c_x,\;c_y)$ & pixel & $(1014.5,\;772.0)^\top$ & $(0.551,\;0.542)^\top$ \\
$\vec{w}$  & - & $(-0.192,\;-2.1,\;25.7)^\top$ & $(0.009,\;0.629,\;13.61)^\top$\\
${^{\mathcal{B}}\vec{r}_{b/n}}$  & mm & $(0.01,\;0.15,\;41.6)^\top$ & $(0.024,\;0.025,\;0.054)^\top$\\
${^{\mathcal{C}}\vec{r}_{n/c}}$  & cm & $(-1.5,\;1.2,\;127.1)^\top$ & $(0.021,\;0.019,\;0.034)^\top$\\
\hline
\end{tabular}
\end{table}

The values of the estimated parameters are reported in Tab. \ref{tab:estimated_uncertainty}. The 1-$\sigma$ uncertainties reported in the table are the square root of the diagonal of the estimated parameters covariance matrix $P_v$. 
In particular, $P_v$ is computed as \cite{hastie2009elements}: 

\begin{equation}
    P_v= \left(J^\top J\right)^{-1}J^\top P_m J \left(J^\top J\right)^{-1}
\end{equation}

where $J$ is the Jacobian after the final iteration and $P_m$ is the measurements covariance matrix. For simplicity, $P_m$ is set equal to:

\begin{equation}
    P_m = \hat{\sigma}^2 \eye{m}
\end{equation}

The variance $\hat{\sigma}^2$ is an unbiased estimator of the true measurement variance computed starting from the total estimation residual $r^2$ \cite{hastie2009elements}: 

\begin{equation}
    \hat{\sigma}^2 = \frac{1}{m-p-1} r^2
\end{equation}

Looking at the values in the table, it is evident that the only variables affected by high uncertainty levels are the second and third polynomial distortion coefficients $w_2$ and $w_3$. This is attributable to the fact that the marker points are not homogeneously distributed in the images but are found only in the central region of the FoV. In that zone, the value of the normalized image coordinates ${^{\scaleto{\mathbb{C}}{3.7pt}}\vec{R}^{n,u}_i}$ is much less than one, therefore the higher-order polynomials have reduced effects.

\subsection{Online attitude estimation results}

Once the calibration is performed, the system is ready for continuous use. The attitude estimation and image processing algorithms have been implemented with C++ in order to guarantee the lowest latency possible. In particular, the latency is defined as the time required to compute an attitude solution once a new image is retrieved.

Figure \ref{fig:PCDF_IP_AE} shows the probability distribution of computational time for the image processing and attitude estimation routines. This data was obtained by running the algorithms on a Raspberry Pi 4. The average latency time is 6.156 ms. As shown, the estimation algorithm takes only 0.5 \% of the total time, which means that large improvements could still be made in reducing the total latency using more efficient image processing schemes. Two distinct peaks can be identified in Fig. \ref{fig:PCDF_IP_AE} corresponding to two different numbers of iterations of the attitude estimation algorithm.  Note that the camera has a maximum acquisition rate of 55.5 Hz, therefore the computation of the solution allocates only about one-third of the time between two consecutive frames (18 ms).

\begin{figure}
\subcaptionbox{Image processing\label{fig:PCDF_IP}}
[.5\textwidth]{\includegraphics[width=0.49\textwidth]{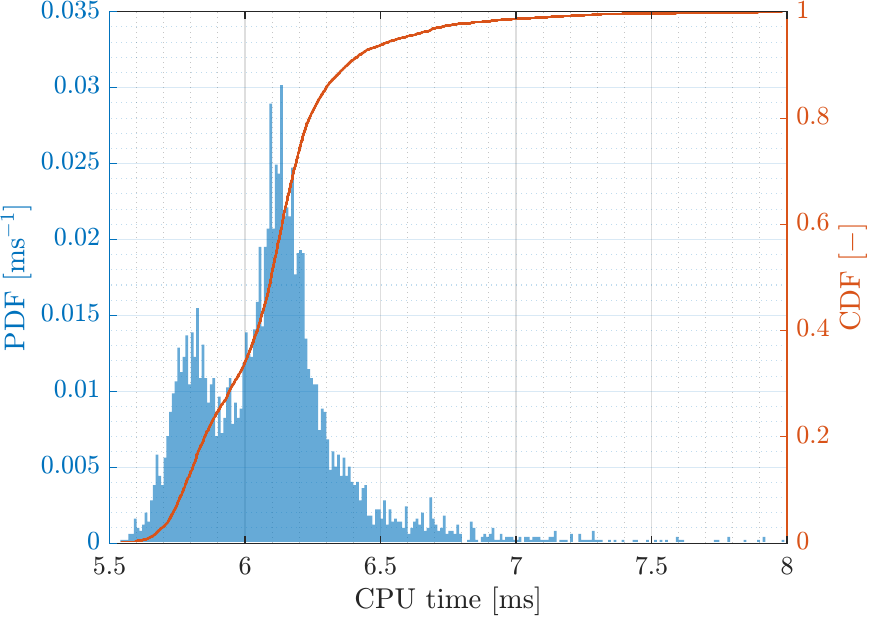}}%
\hfill
\subcaptionbox{Attitude estimation\label{fig:PCDF_AE}}
[.5\textwidth]{\includegraphics[width=0.49\textwidth]{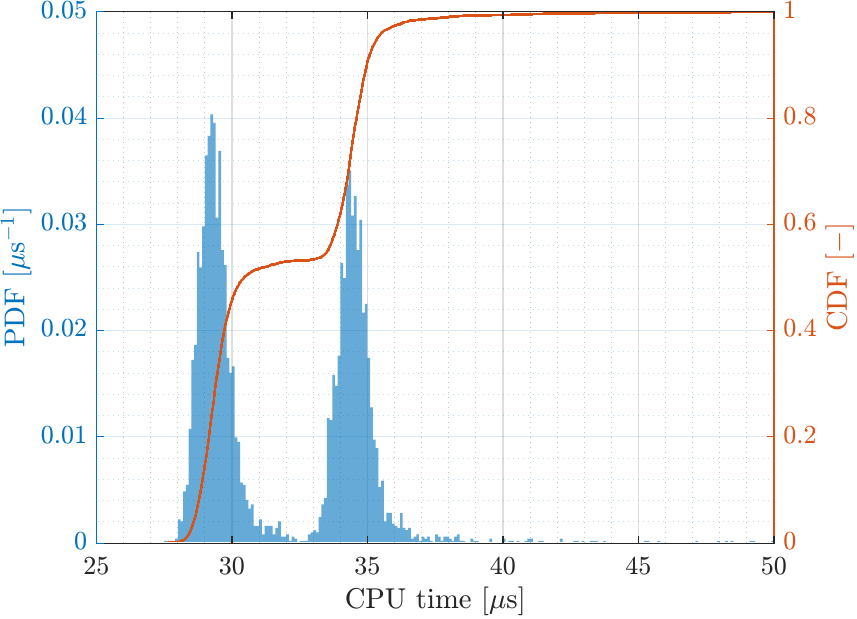}}
\caption{CDF and PDF of execution time for image processing (a) and attitude estimation (b).}
\label{fig:PCDF_IP_AE}
\end{figure}

\begin{figure}
\centering
\subcaptionbox{\label{fig:alpha_hist}}
[.6\textwidth]{\includegraphics[width=0.6\textwidth]{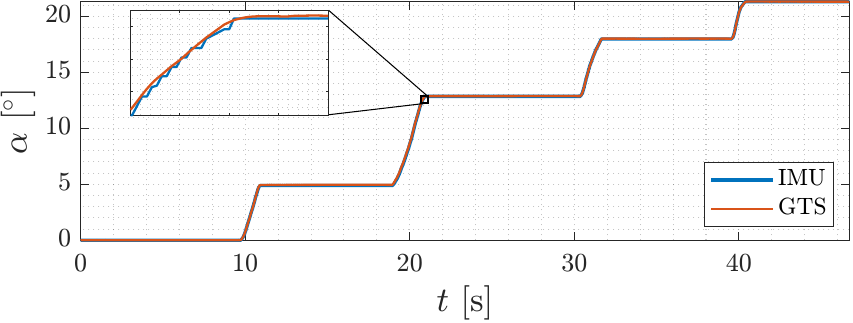}}%
\\
\subcaptionbox{\label{fig:delta_alpha_hist}}
[.6\textwidth]{\includegraphics[width=0.6\textwidth]{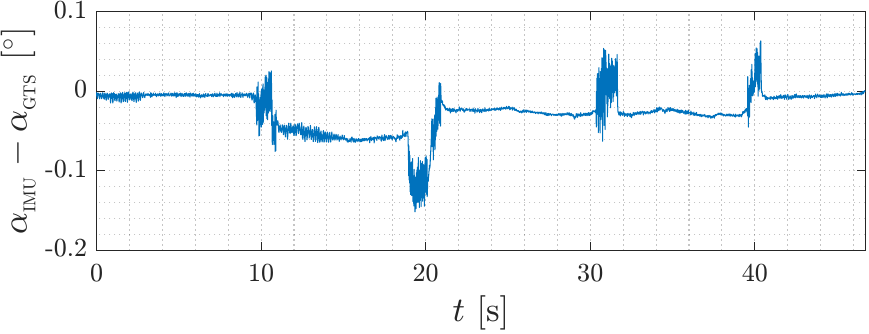}}
\caption{Gravity vector relative rotation computed by IMU and GTS (a) and difference between the two (b).}
\label{fig:_hist}
\end{figure}

The consistency of the online attitude solutions obtained with the GTS is verified using a commercial BNO055 9-axis IMU sensor\footnote{Bosch BNO055 datasheet: \href{https://www.bosch-sensortec.com/media/boschsensortec/downloads/datasheets/bst-bno055-ds000.pdf} {\nolinkurl{https://www.bosch-sensortec.com/media/boschsensortec/downloads/datasheets/bst-bno055-ds000.pdf}}, last accessed on Sept. 2023.}. The IMU collects gyroscope, accelerometer, and magnetometer data to compute the rotation quaternion in an internal sensor-blending filter. Since accelerometer measurements are strongly influenced by centrifugal forces, the test runs are conducted by rotating the air-bearing platform between consecutive static positions. 

In any case, the direct comparison between the IMU and the GTS rotation quaternions is made impossible due to the scarce accuracy of the magnetometer readings and the impossibility of precisely aligning the IMU reference frame with the $\mathcal{B}$ frame. 
For these reasons, a comparison is made by taking into account the relative rotation $\alpha$ of the gravity vector at each time. This approach does not require aligning the frames and reduces the influence of magnetometer data since gravity vector measurements rely mostly on the accelerometer and the gyroscope. Figure \ref{fig:_hist} shows the time history of the rotation angle $\alpha$ as sampled by the IMU at 100 Hz and by the GTS at 55.5 Hz. As can be seen, the measurements from the IMU are coherent with those of the GTS, especially when the platform is static. The maximum difference between the two angle measurements is less than 0.2 degrees and it is experienced during the movements.

\subsection{Sensitivity analysis and expected performance}
Monte Carlo simulations are used to investigate the accuracy and the performance reachable by the methodology with the developed hardware setup in the presence of varying levels of noise and uncertainty. The simulations reproduce the workflow discussed in Section \ref{Methodology workflow} and illustrated in Fig. \ref{fig:workflow_scheme} entirely. The main goal of this analysis is to numerically characterize the current GTS setup to gather a figure of merit that can be compared with the real data.

The geometrical and optical parameters of each simulated system differ from run to run. In particular, their values are selected by randomly perturbing the nominal parameters of the real system with uniformly distributed deviations. The nominal values and the associated maximum allowed deviations are reported in Tab. \ref{tab:simulation_settings}. Additionally, the coordinates of the markers in the body frame ${^{\mathcal{B}}\vec{r}_{i/b}}$ are perturbed with random Gaussian noise having standard deviation $\sigma_p$. This is done to emulate the uncertainty in the knowledge of the actual markers' position due to manufacturing inaccuracies.

\begin{table}[b!]
\centering
\caption{\label{tab:simulation_settings}  Nominal parameters and sampling range for the simulated systems.}
\begin{threeparttable}[h]
\begin{tabular}{lccc}
\hline
Parameter & Unit & Nominal value & Sampling range \\ \hline
$(f_x,\;f_y)$ & pixel & $(3478,\;3478)^\top$ & $\pm$ 50 \\
$(c_x,\;c_y)$ & pixel & $(1024,\;768)^\top$ & $\pm$ 50 \\
$\vec{w}$  & - & $(0,\;0,\; 0)^\top$ & $\pm$ 0.15\\
${^{\mathcal{B}}\vec{r}_{b/n}}$  & cm & $(0,\;0,\;4.2)^\top$ & $\pm$ 1\\
${^{\mathcal{C}}\vec{r}_{n/c}}$  & m & $(0,\;0,\;1.27)^\top$ & $\pm$ 0.05\\
${^{\mathcal{B}}\vec{r}_{s_k/b}}$ & mm & $(0,\;0,\;0)^\top$ & $\pm$  $(5,\;5,\;0)^\top$\\
$\vec{p}_{_{BS_k}}$ & - & $(1,\;0,\;0)^\top$ & $\pm$  $(8.73 \cdot 10^{-3},\;0,\;0)^\top$\tnote{*} \\
\hline
\end{tabular}
\begin{tablenotes}
       \item [*] Equivalent to a rotation of $\pm 1$ degrees around the $\vec{b}_3$ axis.
\end{tablenotes}
\end{threeparttable}
\end{table}

During each run, a virtual system calibration is performed. A total of $N_i = 350$ random attitude poses are generated. In each pose, the location of the virtual centroids ${^{\scaleto{\mathbb{C}}{3.7pt}}\hat{\vec{R}}_i}$ is obtained through the projection model of Section \ref{Camera projection}. A random Gaussian noise with standard deviation $\sigma_i$ is added to each of the ${^{\scaleto{\mathbb{C}}{3.7pt}}\hat{\vec{R}}_i}$ coordinates to simulate inaccuracies in the image processing due to noises. 
After the virtual calibration, 500 additional poses are generated. The attitude of these poses is estimated using the calibration data already computed. The accuracy of the system is verified by comparing the estimated attitudes with the true ones. 

Over 100,000 runs are performed, with $\sigma_i$ values up to 0.3 pixel and $\sigma_p$ up to 0.2 mm. Figure \ref{fig:mean_repr_simulation} shows the average calibration residual $\bar{r}^2$ for each value of $\sigma_i$ and $\sigma_p$. The red contour line highlights the value of $r^2$ obtained after the real hardware calibration, which used the same number of images $N_i$ of the simulations.  
As shown in the figure, the  $r^2$ value obtained with the real hardware is attributable to mean standard deviation errors up to 0.05 mm in the position of the markers and up to 0.12 pixel in the accuracy of the centroids.

\begin{figure}
\centering
\includegraphics[width=0.6\textwidth]{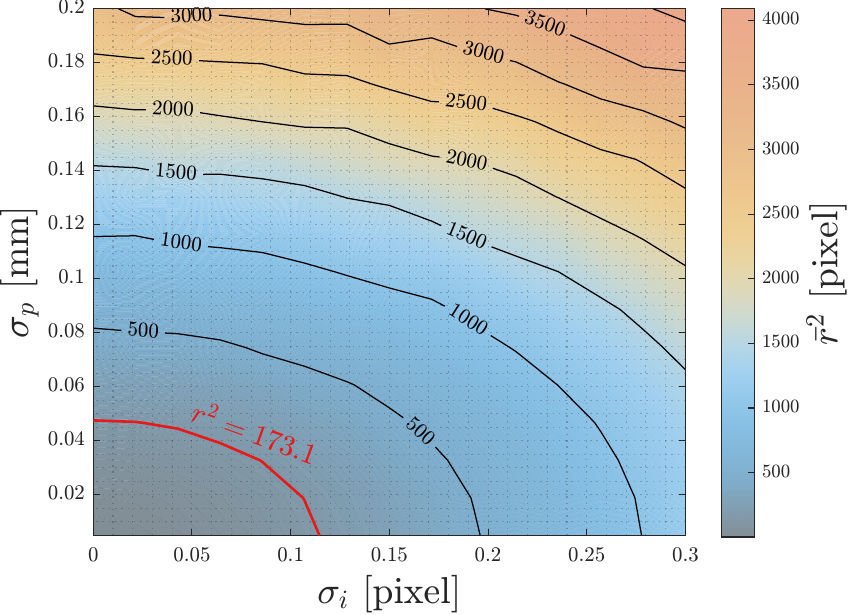}
\caption{Average calibration residual of simulations.}
\label{fig:mean_repr_simulation}
\end{figure}

The same contour line of Fig. \ref{fig:mean_repr_simulation} is highlighted in Fig. \ref{fig:mean_sigma_simulation}, which shows the average standard deviation of the estimated attitude errors for the about-boresight (i.e., yaw) and cross-boresight (i.e., pitch and roll) angles. As the results in the figure suggest, yaw rotations are estimated with greater accuracy than pitch and roll. In particular the average standard deviation of yaw $\Bar{\sigma}_{\rm{yaw}}$ is between 7.5 and 15 arcsec, while the pitch and roll accuracies ($\Bar{\sigma}_{\rm{pitch}}$, $\Bar{\sigma}_{\rm{roll}}$) are lower than 40 arcsec in the vicinity of the real residual contour line.  

\begin{figure}
\subcaptionbox{Yaw\label{fig:mean_sigma_yaw_simulation}}
[.5\textwidth]{\includegraphics[width=0.49\textwidth]{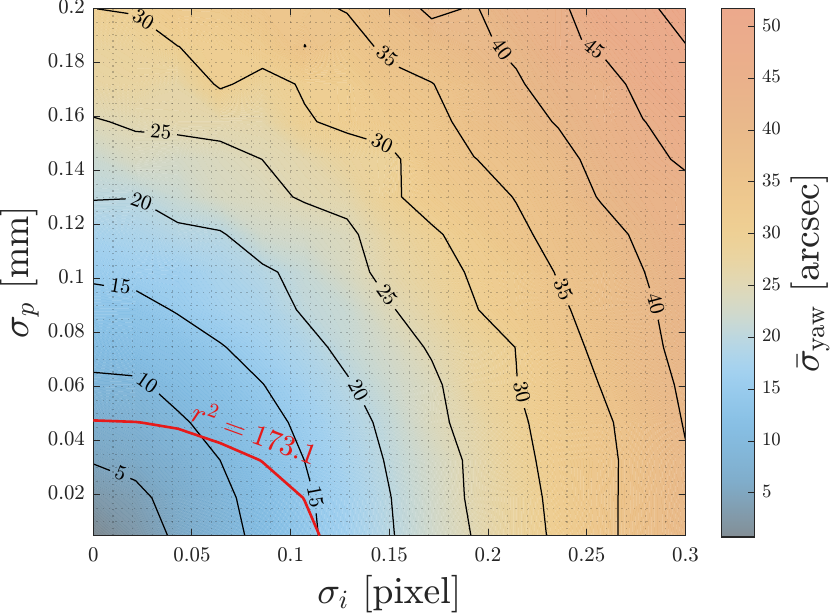}}%
\hfill
\subcaptionbox{Pitch and roll\label{fig:mean_sigma_pitchRoll_simulation}}
[.5\textwidth]{\includegraphics[width=0.49\textwidth]{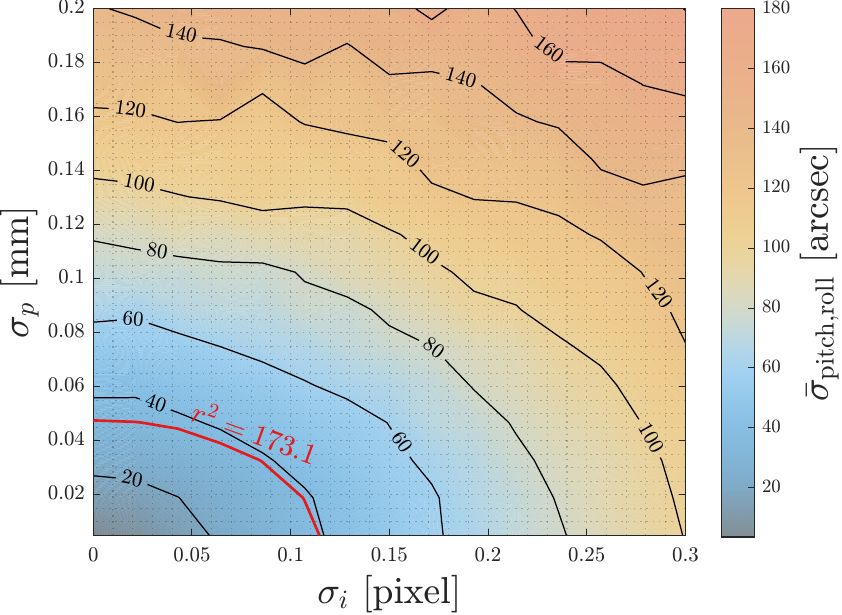}}
\caption{Average standard deviation error of simulations for about- (a) and cross-boresight (b) rotations.}
\label{fig:mean_sigma_simulation}
\end{figure}

Similar results can be observed by comparing the standard deviation error and the total calibration residual of all the simulation runs (see Fig. \ref{fig:sigma_vs_e_simulation}). Evident trends can be identified for $\sigma_{\rm{yaw}}$, $\sigma_{\rm{pitch}}$, and $\sigma_{\rm{roll}}$. In particular, the moving mean line in red confirms the error levels highlighted in Fig. \ref{fig:mean_sigma_simulation}. It is worth noting that all simulations in the vicinity of the real residual line have standard deviations less than 25 arcsec for yaw and 90 arcsec for pitch and roll. 

\begin{figure}
\subcaptionbox{Yaw\label{fig:sigma_yaw_vs_e_simulation}}
[.5\textwidth]{\includegraphics[width=0.49\textwidth]{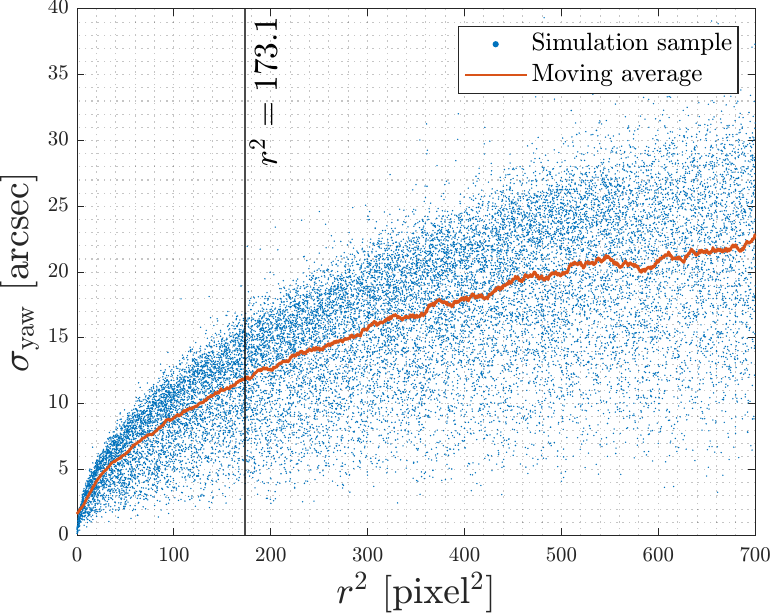}}%
\hfill
\subcaptionbox{Pitch and roll\label{fig:sigma_pitchRoll_vs_e_simulation}}
[.5\textwidth]{\includegraphics[width=0.49\textwidth]{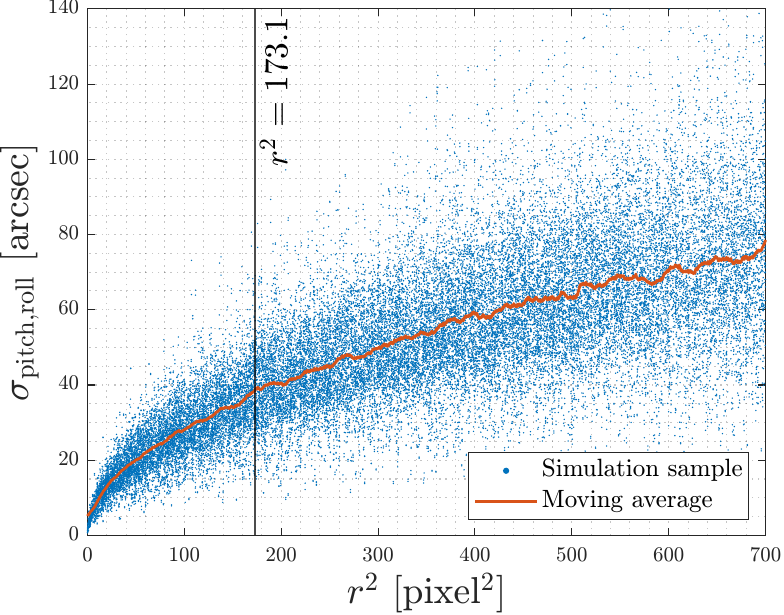}}
\caption{Standard deviation error vs calibration residual for about- (a) and cross-boresight (b) rotations.}
\label{fig:sigma_vs_e_simulation}
\end{figure}

During each simulation run, the attitude is also computed through the MATLAB P3P algorithm \cite{gao2003complete} and the Infinitesimal Plane-based Pose Estimation (IPPE) P$n$P algorithm \cite{Collins2014Infinitesimal}. Note that these algorithms cannot work with multiple independent pattern sets, therefore they are made aware of the exact relative position and orientation of the pattern sets.

Figure \ref{fig:sigma_comparison_ratio} shows the ratio between the 1-$\sigma$ accuracy of the P3P and P$n$P algorithms and the one of the developed method in each simulation run. Results are compared against the total residual after calibration of each run. 
In 100 \% of the tested cases, the accuracy ratio is higher than one. In particular, for pitch and roll, the developed method is on average 7.5 times more accurate than IPPE and 17 times more accurate than MATLAB's P3P algorithm in simulations with calibration residuals similar to the one of the real system. For yaw solutions, instead, the average accuracy ratio is 2.5 against IPPE and 9.5 against the P3P.  

It is important to note that in these simulations the P3P and P$n$P algorithms have the exact knowledge not only of relative geometry between the patterns but also of camera intrinsic parameters. In a real scenario, those parameters would be affected by estimation errors which could have a relevant impact on the obtainable accuracy.
The accuracy ratios discussed are intended as a worst-case indication of the performance improvement brought by the developed method against state-of-the-art P3P and P$n$P implementations in optimal conditions. Therefore, even higher accuracy ratios are expected when calibration errors are considered.
This goes to show the versatility of the proposed methodology, which retains better performances while working with multiple independent patterns after performing an estimation of the parameters of the system.

\begin{figure}
\subcaptionbox{Yaw\label{fig:sigma_comparison_ratio_yaw}}
[.5\textwidth]{\includegraphics[width=0.49\textwidth]{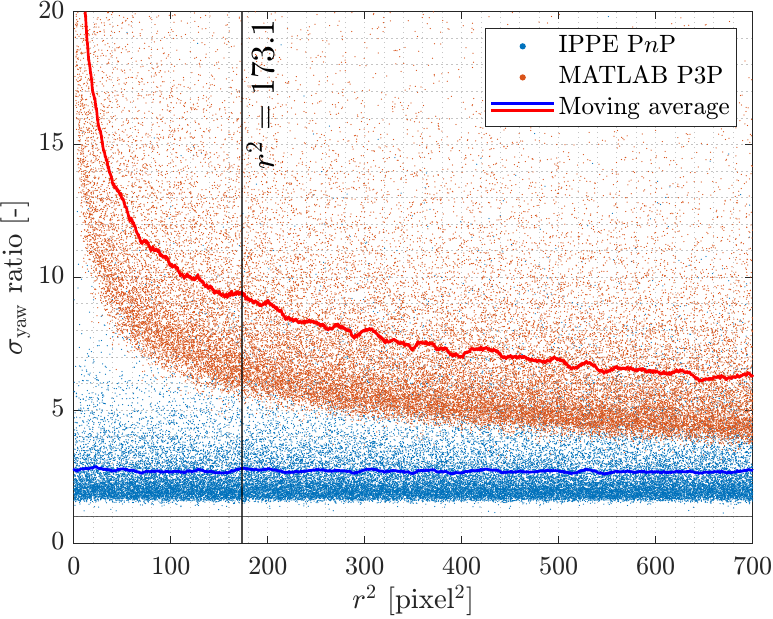}}%
\hfill
\subcaptionbox{Pitch and roll\label{fig:sigma_comparison_ratio_pitch_roll}}
[.5\textwidth]{\includegraphics[width=0.49\textwidth]{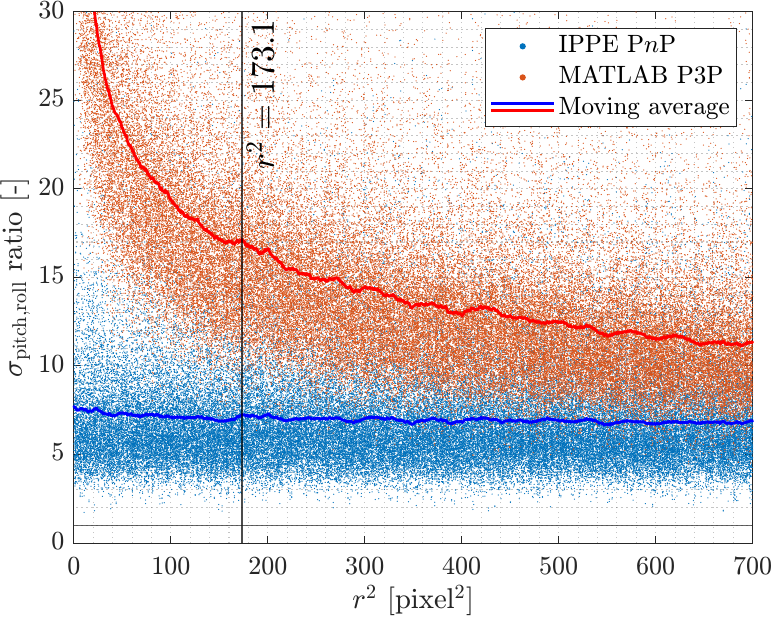}}
\caption{Ratio between accuracy of other algorithms and developed method vs calibration residual for about- (a) and cross-boresight (b) rotations.}
\label{fig:sigma_comparison_ratio}
\end{figure}

\section{Conclusion}
\label{sec:4}

This paper presents a vision-based method to precisely compute the attitude of 3-DoF rotational air-bearing platforms. The proposed methodology is divided into two phases. During an initial calibration phase, the system's characteristics are computed through a batch estimation of the geometrical and optical parameters of the system. In the second phase, the platform's attitude is determined with an iterative approach.
A prototype hardware setup composed of a fiducial pattern of LEDs is designed and installed onto a real air-bearing platform.
Using a Raspberry Pi 4 as computing hardware it is possible to retrieve attitude solutions with an average latency of 6.2 ms at a 55 Hz refresh rate. 
Finally, a Monte Carlo simulation is performed to investigate the accuracy reachable by the assembled system. The simulations indicate an average 1-$\sigma$ accuracy of 12 arcsec for yaw and 37 arcsec for pitch and roll. The results of the method are compared with two state-of-the-art Perspective-$n$-Point implementations, showing a considerable improvement in the accuracy up to one order in magnitude.
The presented solution constitutes a low-cost, versatile, and effective alternative to commercial motion-capture systems, that can be easily adapted to different hardware architectures.

\section*{Appendices}
\subsection*{Appendix A - Attitude estimation Jacobian} 
\label{sec:appA}

The Jacobian $J_{p}$ used in the attitude estimation algorithm is composed of all the partial derivatives of the position of each marker ${^{\scaleto{\mathbb{C}}{3.7pt}}\vec{R}_i}$ in the camera frame with respect to the attitude parameters $\vec{p}_{_{NB}}$:

\begin{equation}
    J_{p} = \begin{bmatrix}
        \dfrac{\partial \; {^{\scaleto{\mathbb{C}}{3.7pt}}\vec{R}_1}}{\partial \vec{p}_{_{NB}}} \\ \dfrac{\partial \;{^{\scaleto{\mathbb{C}}{3.7pt}}\vec{R}_2}}{\partial \vec{p}_{_{NB}}}  \\
        \vdots  \\
        \dfrac{\partial \; {^{\scaleto{\mathbb{C}}{3.7pt}}\vec{R}_{N_m}}}{\partial \vec{p}_{_{NB}}}
    \end{bmatrix}
\end{equation}

Each partial derivative can be expanded as: 

\begin{equation}
    \dfrac{ \partial {\, {^{\scaleto{\mathbb{C}}{3.7pt}}\vec{R}_i} } }{ \partial \, \vec{p}_{_{NB}}} = \dfrac{ \partial \, { {^{\scaleto{\mathbb{C}}{3.7pt}}\vec{R}_i} } }{ \partial \, {^{\scaleto{\mathbb{C}}{3.7pt}}\vec{R}^{n}_i}} 
    \dfrac { \partial \, {^{\scaleto{\mathbb{C}}{3.7pt}}\vec{R}^{n}_i}} {\partial \,{^{\scaleto{\mathbb{C}}{3.7pt}}\vec{R}^{n,u}_i}} \dfrac { \partial \, {^{\scaleto{\mathbb{C}}{3.7pt}}\vec{R}^{n,u}_i}} {\partial \,{^{\scaleto{\mathbb{C}}{3.7pt}}\vec{R}^{u}_i}} \dfrac {{\partial \, ^{\scaleto{\mathbb{C}}{3.7pt}}\vec{R}^{u}_i}} {\partial \,{^{\scaleto{\mathbb{C}}{3.7pt}}\vec{R}^{h}_i}}
    \dfrac {\partial \, {^{\scaleto{\mathbb{C}}{3.7pt}}\vec{R}^{h}_i}} {\partial \,{^{\mathcal{C}}\vec{r}_{i/c}}}
    \dfrac {\partial \,{^{\mathcal{C}}\vec{r}_{i/c}}}
    { \partial \, { {^{\mathcal{N}}\vec{r}_{i/n}}}}
    \dfrac {\partial \, { {^{\mathcal{N}}\vec{r}_{i/n}}}} {\partial {\vec{q}_{_{NB}}}}
    \dfrac {\partial {\vec{q}_{_{NB}}}} {\partial {\vec{p}_{_{NB}}}}
\end{equation}

where: 

\begin{gather} 
    \dfrac{ \partial \, { {^{\scaleto{\mathbb{C}}{3.7pt}}\vec{R}_i} } }{ \partial \, {^{\scaleto{\mathbb{C}}{3.7pt}}\vec{R}^{n}_i}} = \begin{bmatrix}
        f_x & 0 \\ 0& f_y
    \end{bmatrix} \\
     \dfrac { \partial \, {^{\scaleto{\mathbb{C}}{3.7pt}}\vec{R}^{n}_i}} {\partial \,{^{\scaleto{\mathbb{C}}{3.7pt}}\vec{R}^{n,u}_i}} = \left(1 + w_1 \rho_i^2 + w_2 \rho_i^4 + w_3 \rho_i^6 \right) \eye{2} + \left(2 w_1 + 4 w_2 \rho_i^2 + 6 w_3 \rho_i^4 \right) {^{\scaleto{\mathbb{C}}{3.7pt}}\vec{R}^{n,u}_i} {^{\scaleto{\mathbb{C}}{3.7pt}}\vec{R}^{n,u}_i}^\top \\
     \dfrac { \partial \, {^{\scaleto{\mathbb{C}}{3.7pt}}\vec{R}^{n,u}_i}} {\partial \,{^{\scaleto{\mathbb{C}}{3.7pt}}\vec{R}^{u}_i}} =  \begin{bmatrix}
        \frac {1} {f_x} & 0 \\ 0& \frac {1} {f_y} 
    \end{bmatrix} \\
     \dfrac {{\partial \, ^{\scaleto{\mathbb{C}}{3.7pt}}\vec{R}^{u}_i}} {\partial \,{^{\scaleto{\mathbb{C}}{3.7pt}}\vec{R}^{h}_i}} = \dfrac{1}{{^{\scaleto{\mathbb{C}}{3.7pt}}\vec{R}^{h}_i} \cdot {\vec{C}_3}}\begin{bmatrix}
         1 & 0 & {-{^{\scaleto{\mathbb{C}}{3.7pt}}\vec{R}^{h}_i} \cdot {\vec{C}_1}} \\
         0 & 1 & {-{^{\scaleto{\mathbb{C}}{3.7pt}}\vec{R}^{h}_i} \cdot {\vec{C}_2}}  \end{bmatrix} \\
         \dfrac {\partial \, {^{\scaleto{\mathbb{C}}{3.7pt}}\vec{R}^{h}_i}} {\partial \,{^{\mathcal{C}}\vec{r}_{i/c}}} = K \\
         \dfrac {\partial \,{^{\mathcal{C}}\vec{r}_{i/c}}}
    { \partial \, { {^{\mathcal{N}}\vec{r}_{i/n}}}} = [CN] 
\end{gather}

The derivative of a rotated vector $\vec{g}=R(\vec{q}) \vec{u}$ with respect to the rotation quaternion $\vec{q} = (q_w,\;\vec{q}_v)^\top$ can be computed as \cite{sola2017quaternion}:   

\begin{equation}
\label{eq:Rq_dq}
\dfrac{\partial \vec{g} }{\partial \vec{q}} =    \dfrac{\partial R(\vec{q}) \vec{u} }{\partial \vec{q}} = 2 \begin{bmatrix}
        q_w \vec{u} + \vec{u} \times \vec{q}_v \;, & \left(\vec{q}_v^\top \cdot \vec{u}\right) \eye{3} - \vec{u}\vec{q}_v^\top + \vec{q}_v\vec{u}^\top + q_w [\vec{u}^\wedge]
    \end{bmatrix}
\end{equation}

Where $[\vec{u}^\wedge]$ is the skew-symmetric matrix of the vector $\vec{u}$. Looking at Eq. \eqref{eq:r_nin}, it is evident that $\dfrac {\partial \, { {^{\mathcal{N}}\vec{r}_{i/n}}}} {\partial {\vec{q}_{_{NB}}}}$ can be computed by substituting $\vec{u} = {^{\mathcal{B}}\vec{r}_{i/b}}+{^{\mathcal{B}}\vec{r}_{b/n}}$ in Eq. \eqref{eq:Rq_dq}.

The derivative of a unitary quaternion vector $\vec{q}$ with respect to its associated vector of parameters $\vec{p} = (p_1,\; p_2,\; p_3)^\top$ is \cite{terzakis2014quaternion}: 

\begin{equation}
\label{eq:p_derivative}
    \dfrac{\partial \vec{q} }{\partial \vec{p}} = -\dfrac{4}{\left(\alpha^2 +1\right)^2} \begin{bmatrix}
        p_1^2 - \frac{\alpha^2+1}{2} & p_1 p_2 & p_1 p_3 \\
        p_1 p_2 & p_2^2 - \frac{\alpha^2+1}{2} & p_2 p_3 \\
        p_1 p_3 & p_2 p_3 & p_3^2 - \frac{\alpha^2+1}{2} \\
        p_1 & p_2 & p_3        
    \end{bmatrix}
\end{equation}

\subsection*{Appendix B - System calibration Jacobian}
\label{sec:appB}

The Jacobian matrix $J$ used during the system calibration procedure is widely sparse. In fact, the derivative of the component ${\boldsymbol{\varepsilon}_m}$ of reprojection error vector $\boldsymbol{\eta}$ with respect to the attitude parameters $\vec{p}_{_{NB_j}}$ of the $j$-th image is:

\begin{equation}
    \dfrac{ \partial{\boldsymbol{\varepsilon}_l}}{ \partial \vec{p}_{_{NB_j}}} = \begin{cases}
        J_{p_j} &\text{if $l=j$} \\
        \zeros{2N_m}{3} & \text{if $l \neq j$}
    \end{cases}
\end{equation}

The derivative of pixel coordinates ${^{\scaleto{\mathbb{C}}{3.7pt}}\vec{R}_i}$ with respect to the vector of camera parameters $\vec{k}$ is:

\begin{equation} 
    \dfrac{ \partial \, { {^{\scaleto{\mathbb{C}}{3.7pt}}\vec{R}_i} } } { \partial \, \vec{k} } =
    \begin{bmatrix}
         {{^{\scaleto{\mathbb{C}}{3.7pt}}\vec{R}^{n}_i} \cdot {\vec{C}_1}} & 0 & 1 & 0 \\
         0 & {{^{\scaleto{\mathbb{C}}{3.7pt}}\vec{R}^{n}_i} \cdot {\vec{C}_2}} & 0 & 1 \end{bmatrix}  + \dfrac{ \partial \, { {^{\scaleto{\mathbb{C}}{3.7pt}}\vec{R}_i} } }{ \partial \, {^{\scaleto{\mathbb{C}}{3.7pt}}\vec{R}^{n}_i}} 
    \dfrac { \partial \, {^{\scaleto{\mathbb{C}}{3.7pt}}\vec{R}^{n}_i}} {\partial \,{^{\scaleto{\mathbb{C}}{3.7pt}}\vec{R}^{n,u}_i}} \left( \dfrac {\partial \,{^{\scaleto{\mathbb{C}}{3.7pt}}\vec{R}^{n,u}_i}} { \partial \, \vec{k} } + \dfrac { \partial \, {^{\scaleto{\mathbb{C}}{3.7pt}}\vec{R}^{n,u}_i}} {\partial \,{^{\scaleto{\mathbb{C}}{3.7pt}}\vec{R}^{u}_i}} \dfrac {{\partial \, ^{\scaleto{\mathbb{C}}{3.7pt}}\vec{R}^{u}_i}} {\partial \,{^{\scaleto{\mathbb{C}}{3.7pt}}\vec{R}^{h}_i}}
    \dfrac {\partial \, {^{\scaleto{\mathbb{C}}{3.7pt}}\vec{R}^{h}_i}} { \partial \, \vec{k} } \right)
\end{equation}

where: 

\begin{gather}
    \dfrac {\partial \,{^{\scaleto{\mathbb{C}}{3.7pt}}\vec{R}^{n,u}_i}} { \partial \, \vec{k} } = - \begin{bmatrix}
         \dfrac{{{^{\scaleto{\mathbb{C}}{3.7pt}}\vec{R}^{u}_i} \cdot {\vec{C}_1}} - c_x}{f_x^2} & 0 & \dfrac{1}{f_x} & 0 \\
         0 & \dfrac{{{^{\scaleto{\mathbb{C}}{3.7pt}}\vec{R}^{u}_i} \cdot {\vec{C}_2}} - c_y}{f_y^2} & 0 & \dfrac{1}{f_y} \end{bmatrix}
    \\ \dfrac {\partial \, {^{\scaleto{\mathbb{C}}{3.7pt}}\vec{R}^{h}_i}} { \partial \, \vec{k} } = \begin{bmatrix}
        {^{\mathcal{C}}\vec{r}_{i/c}} \cdot \vec{c}_1 & 0 & {^{\mathcal{C}}\vec{r}_{i/c}} \cdot \vec{c}_3 &0 \\ 
        0 & {^{\mathcal{C}}\vec{r}_{i/c}} \cdot \vec{c}_2 & 0 & {^{\mathcal{C}}\vec{r}_{i/c}} \cdot \vec{c}_3 \\
        0 & 0 & 0 & 0
    \end{bmatrix}  
\end{gather}

The derivative of ${^{\scaleto{\mathbb{C}}{3.7pt}}\vec{R}_i}$ with respect to the vector of radial distortion coefficients $\vec{w}$ is: 

\begin{equation}
    \dfrac{ \partial \, { {^{\scaleto{\mathbb{C}}{3.7pt}}\vec{R}_i} } } { \partial \, \vec{w} } = 
    \dfrac{ \partial \, { {^{\scaleto{\mathbb{C}}{3.7pt}}\vec{R}_i} } }{ \partial \, {^{\scaleto{\mathbb{C}}{3.7pt}}\vec{R}_i^{n}}} \dfrac { \partial \, {^{\scaleto{\mathbb{C}}{3.7pt}}\vec{R}_i^{n}}} { \partial \, \vec{w} }
\end{equation}

where: 
\begin{equation}
    \dfrac { \partial \, {^{\scaleto{\mathbb{C}}{3.7pt}}\vec{R}^{n}_i}} { \partial \, \vec{w} } = {^{\scaleto{\mathbb{C}}{3.7pt}}\vec{R}^{n,u}_i}\left( 
        \rho_i^2, \; \rho_i^4, \; \rho_i^6\right)
\end{equation}

The derivative of ${^{\scaleto{\mathbb{C}}{3.7pt}}\vec{R}_i}$ with respect to the position of the center of rotation in the body frame ${^{\mathcal{B}}\vec{r}_{b/n}}$ can be expanded as: 

\begin{equation}
    \dfrac{ \partial {\, {^{\scaleto{\mathbb{C}}{3.7pt}}\vec{R}_i} } }{ \partial \, {^{\mathcal{B}}\vec{r}_{b/n}}} = \dfrac{ \partial \, { {^{\scaleto{\mathbb{C}}{3.7pt}}\vec{R}_i} } }{ \partial \, {^{\scaleto{\mathbb{C}}{3.7pt}}\vec{R}^{n}_i}} 
    \dfrac { \partial \, {^{\scaleto{\mathbb{C}}{3.7pt}}\vec{R}^{n}_i}} {\partial \,{^{\scaleto{\mathbb{C}}{3.7pt}}\vec{R}^{n,u}_i}} \dfrac { \partial \, {^{\scaleto{\mathbb{C}}{3.7pt}}\vec{R}^{n,u}_i}} {\partial \,{^{\scaleto{\mathbb{C}}{3.7pt}}\vec{R}^{u}_i}} \dfrac {{\partial \, ^{\scaleto{\mathbb{C}}{3.7pt}}\vec{R}^{u}_i}} {\partial \,{^{\scaleto{\mathbb{C}}{3.7pt}}\vec{R}^{h}_i}}
    \dfrac {\partial \, {^{\scaleto{\mathbb{C}}{3.7pt}}\vec{R}^{h}_i}} {\partial \,{^{\mathcal{C}}\vec{r}_{i/c}}}
    \dfrac {\partial \,{^{\mathcal{C}}\vec{r}_{i/c}}}
    { \partial \, { {^{\mathcal{N}}\vec{r}_{i/n}}}}
    \dfrac {\partial \, { {^{\mathcal{N}}\vec{r}_{i/n}}}} {\partial \, {^{\mathcal{B}}\vec{r}_{b/n}}}
\end{equation}

In which the only derivative not yet shown is: 

\begin{equation}
    \dfrac {\partial \, { {^{\mathcal{N}}\vec{r}_{i/n}}}} {\partial \, {^{\mathcal{B}}\vec{r}_{b/n}}} = [NB]_j
\end{equation}

Similarly, the derivative of ${^{\scaleto{\mathbb{C}}{3.7pt}}\vec{R}_i}$ with respect to the position of the center of rotation in the camera frame ${^{\mathcal{C}}\vec{r}_{n/c}}$ can be computed as: 

\begin{equation}
    \dfrac{ \partial {\, {^{\scaleto{\mathbb{C}}{3.7pt}}\vec{R}_i} } }{ \partial \, {^{\mathcal{C}}\vec{r}_{n/c}}} = \dfrac{ \partial \, { {^{\scaleto{\mathbb{C}}{3.7pt}}\vec{R}_i} } }{ \partial \, {^{\scaleto{\mathbb{C}}{3.7pt}}\vec{R}^{n}_i}} 
    \dfrac { \partial \, {^{\scaleto{\mathbb{C}}{3.7pt}}\vec{R}^{n}_i}} {\partial \,{^{\scaleto{\mathbb{C}}{3.7pt}}\vec{R}^{n,u}_i}} \dfrac { \partial \, {^{\scaleto{\mathbb{C}}{3.7pt}}\vec{R}^{n,u}_i}} {\partial \,{^{\scaleto{\mathbb{C}}{3.7pt}}\vec{R}^{u}_i}} \dfrac {{\partial \, ^{\scaleto{\mathbb{C}}{3.7pt}}\vec{R}^{u}_i}} {\partial \,{^{\scaleto{\mathbb{C}}{3.7pt}}\vec{R}^{h}_i}}
    \dfrac {\partial \, {^{\scaleto{\mathbb{C}}{3.7pt}}\vec{R}^{h}_i}} {\partial \,{^{\mathcal{C}}\vec{r}_{i/c}}}
    \dfrac {\partial \,{^{\mathcal{C}}\vec{r}_{i/c}}} {\partial \,{^{\mathcal{C}}\vec{r}_{n/c}}}
\end{equation}

Where: 

\begin{equation}
    \dfrac {\partial \,{^{\mathcal{C}}\vec{r}_{i/c}}} {\partial \,{^{\mathcal{C}}\vec{r}_{n/c}}} = \eye{3}
\end{equation}

If additional patterns are present, the derivative of ${^{\scaleto{\mathbb{C}}{3.7pt}}\vec{R}_i}$ with respect to the position of the origin of each frame relative to the body frame ${^{\mathcal{B}}\vec{r}_{s_k/b}}$ is: 

\begin{equation}
    \dfrac{ \partial {\, {^{\scaleto{\mathbb{C}}{3.7pt}}\vec{R}_i} } }{ \partial \, {^{\mathcal{B}}\vec{r}_{s_k/b}}} = \dfrac{ \partial \, { {^{\scaleto{\mathbb{C}}{3.7pt}}\vec{R}_i} } }{ \partial \, {^{\scaleto{\mathbb{C}}{3.7pt}}\vec{R}^{n}_i}} 
    \dfrac { \partial \, {^{\scaleto{\mathbb{C}}{3.7pt}}\vec{R}^{n}_i}} {\partial \,{^{\scaleto{\mathbb{C}}{3.7pt}}\vec{R}^{n,u}_i}} \dfrac { \partial \, {^{\scaleto{\mathbb{C}}{3.7pt}}\vec{R}^{n,u}_i}} {\partial \,{^{\scaleto{\mathbb{C}}{3.7pt}}\vec{R}^{u}_i}} \dfrac {{\partial \, ^{\scaleto{\mathbb{C}}{3.7pt}}\vec{R}^{u}_i}} {\partial \,{^{\scaleto{\mathbb{C}}{3.7pt}}\vec{R}^{h}_i}}
    \dfrac {\partial \, {^{\scaleto{\mathbb{C}}{3.7pt}}\vec{R}^{h}_i}} {\partial \,{^{\mathcal{C}}\vec{r}_{i/c}}}
    \dfrac {\partial \,{^{\mathcal{C}}\vec{r}_{i/c}}} {\partial \,{^{\mathcal{N}}\vec{r}_{i/n}}}
    \dfrac {\partial \,{^{\mathcal{N}}\vec{r}_{i/n}}} {\partial \,{^{\mathcal{B}}\vec{r}_{i/b}}}
    \dfrac {\partial \,{^{\mathcal{B}}\vec{r}_{i/b}}} {\partial \, {^{\mathcal{B}}\vec{r}_{s_k/b}}}
\end{equation}

Where: 

\begin{gather}
    \dfrac {\partial \,{^{\mathcal{N}}\vec{r}_{i/n}}} {\partial \,{^{\mathcal{B}}\vec{r}_{i/b}}} = [NB]_j \\
    \dfrac {\partial \,{^{\mathcal{B}}\vec{r}_{i/b}}} {\partial \, {^{\mathcal{B}}\vec{r}_{s_k/b}}} = \begin{cases}
        [BS_k] & \text{if marker $i \in$  pattern $k$}\\
        \zeros{3}{3} & \text{otherwise}
    \end{cases}
\end{gather}

At the same time, the derivative of ${^{\scaleto{\mathbb{C}}{3.7pt}}\vec{R}_i}$ with respect to the rotation parameters of the additional patterns $\vec{p}_{_{BS_k}}$ is: 

\begin{equation}
    \dfrac{ \partial {\, {^{\scaleto{\mathbb{C}}{3.7pt}}\vec{R}_i} } }{ \partial \, {^{\mathcal{B}}\vec{r}_{s_k/b}}} = \dfrac{ \partial \, { {^{\scaleto{\mathbb{C}}{3.7pt}}\vec{R}_i} } }{ \partial \, {^{\scaleto{\mathbb{C}}{3.7pt}}\vec{R}^{n}_i}} 
    \dfrac { \partial \, {^{\scaleto{\mathbb{C}}{3.7pt}}\vec{R}^{n}_i}} {\partial \,{^{\scaleto{\mathbb{C}}{3.7pt}}\vec{R}^{n,u}_i}} \dfrac { \partial \, {^{\scaleto{\mathbb{C}}{3.7pt}}\vec{R}^{n,u}_i}} {\partial \,{^{\scaleto{\mathbb{C}}{3.7pt}}\vec{R}^{u}_i}} \dfrac {{\partial \, ^{\scaleto{\mathbb{C}}{3.7pt}}\vec{R}^{u}_i}} {\partial \,{^{\scaleto{\mathbb{C}}{3.7pt}}\vec{R}^{h}_i}}
    \dfrac {\partial \, {^{\scaleto{\mathbb{C}}{3.7pt}}\vec{R}^{h}_i}} {\partial \,{^{\mathcal{C}}\vec{r}_{i/c}}}
    \dfrac {\partial \,{^{\mathcal{C}}\vec{r}_{i/c}}} {\partial \,{^{\mathcal{N}}\vec{r}_{i/n}}}
    \dfrac {\partial \,{^{\mathcal{N}}\vec{r}_{i/n}}} {\partial \,{^{\mathcal{B}}\vec{r}_{i/b}}}
    \dfrac {\partial \,{^{\mathcal{B}}\vec{r}_{i/b}}} {\partial \vec{q}_{_{BS_k}}}
    \dfrac {\partial \vec{q}_{_{BS_k}}} {\partial \vec{p}_{_{BS_k}}}
\end{equation}

Note that, this derivative is also null if the marker point $i$ does not belong to the additional pattern $k$. The missing derivatives $\dfrac {\partial \,{^{\mathcal{B}}\vec{r}_{i/b}}} {\partial \vec{q}_{_{BS_k}}}$ and $
    \dfrac {\partial \vec{q}_{_{BS_k}}} {\partial \vec{p}_{_{BS_k}}}$ can be computed using Eq. \eqref{eq:p_derivative} and by substituting $\vec{u} = {^{{\mathcal{S}^k}}\vec{r}_{i/s_k}}$ in Eq. \eqref{eq:Rq_dq}.

\section*{Acknowledgments}
This project has received funding from the European Research Council (ERC) under the European Union’s Horizon 2020 research and innovation programme (grant agreement No. 864697).

\bibliography{sample}

\end{document}